\documentclass[10pt,twocolumn,letterpaper]{article}

\usepackage{iccv}              
\usepackage{algorithm}
\usepackage{algorithmic}
\usepackage{booktabs}

\definecolor{iccvblue}{rgb}{0.21,0.49,0.74}
\usepackage[pagebackref,breaklinks,colorlinks,allcolors=iccvblue]{hyperref}

\def\paperID{13468} 
\def\confName{ICCV}
\def\confYear{2025}
\usepackage{subcaption} 
\usepackage{amsmath, amssymb} 
\usepackage{amsthm}
\newtheorem{theorem}{Theorem}
\newtheorem{definition}{Definition}
\newtheorem{assumption}{Assumption}
\newtheorem{lemma}{Lemma}
\usepackage{multirow}
\usepackage{mdframed}

\newcommand{\lime}{{\sc lime}\xspace}
\newcommand{\shap}{{\sc shap}\xspace}
\newcommand{\fastshap}{{\sc f}ast{\sc shap}\xspace}
\newcommand{\rise}{{\sc rise}\xspace}
\newcommand{\yolo}{{\sc yolo}\xspace}
\newcommand{\rtdetr}{{\sc rt-detr}\xspace}
\newcommand{\drise}{{\sc d-rise}\xspace}

\newcommand{\rex}{{\sc r}e{\sc x}\xspace}

\newcommand{\xai}{XAI\xspace}
\newcommand{\commentout}[1]{}
\newcommand{\incx}{\textsc{IncX}\xspace}
\newcommand{\fastrcnn}{\textsc{f}\textnormal{aster}\,\textsc{r}-\textsc{cnn}\xspace}
\newcommand{\rcnn}{\textsc{r}-\textsc{cnn}\xspace}
\newcommand{\frcnn}{\textsc{f}\textnormal{ast}\,\textsc{r}-\textsc{cnn}\xspace}
\newcommand{\fsod}{\textsc{fsod}\xspace}

\newenvironment{proofsketch}{%
  \proof}{\endproof}

\usepackage[capitalize,noabbrev]{cleveref}

\usepackage{tikz}
\usetikzlibrary{positioning,quotes}
\usetikzlibrary{backgrounds}

\usepackage{subcaption}

\usepackage{tabularx}
\newcolumntype{B}{p{3.5cm}}
\newcolumntype{s}{>{\raggedleft\arraybackslash $}p{1.3cm}<{$}}
\newcolumntype{q}{>{\raggedleft}p{1.7cm}<}

\title{Real-Time Incremental Explanations for Object Detectors\\ in Autonomous Driving}

\author{
    Santiago Calder\'{o}n-Pe\~{n}a,
    Hana Chockler,
    David A. Kelly \\
    King's College London \\
    London, United Kingdom \\
    {\tt\small \{santiago.calderon, hana.chockler, david.a.kelly\}@kcl.ac.uk}
}

\begin{document}
\maketitle

\begin{abstract}
Object detectors are widely used in safety-critical real-time applications such as autonomous driving. 
Explainability is especially important for safety-critical applications, and due to the variety of object detectors and
their often proprietary nature, black-box explainability tools are needed.
However, existing black-box explainability tools for AI models rely on multiple model calls, rendering them impractical 
for real-time use. 

In this paper, we introduce \incx, an algorithm and a tool for real-time black-box explainability for object detectors.
The algorithm is based on linear transformations of saliency maps, producing sufficient explanations. 
We evaluate our implementation on four widely used video datasets of autonomous driving and demonstrate that \incx's explanations
are comparable in quality to the state-of-the-art and are computed two orders of magnitude faster than the state-of-the-art,
making them usable in real time.
\end{abstract}

\section{Introduction}

\setlength{\fboxsep}{0pt}
\setlength{\fboxrule}{0.4pt}

\begin{figure*}[t]
    \centering
    \begin{tikzpicture}[
        label distance=2mm,
        every edge/.style = {draw=red, ->, very thick},
        every edge quotes/.style = {auto, font=\footnotesize,text=black, fill=none, sloped}]
    ]
        \node[inner sep=0pt] (orig) at (-2.3, 0) {Frames};
        \node[inner sep=0pt, label=above:$t_0$] (frame1) at (0,0) {\fbox{\includegraphics[scale=0.23]{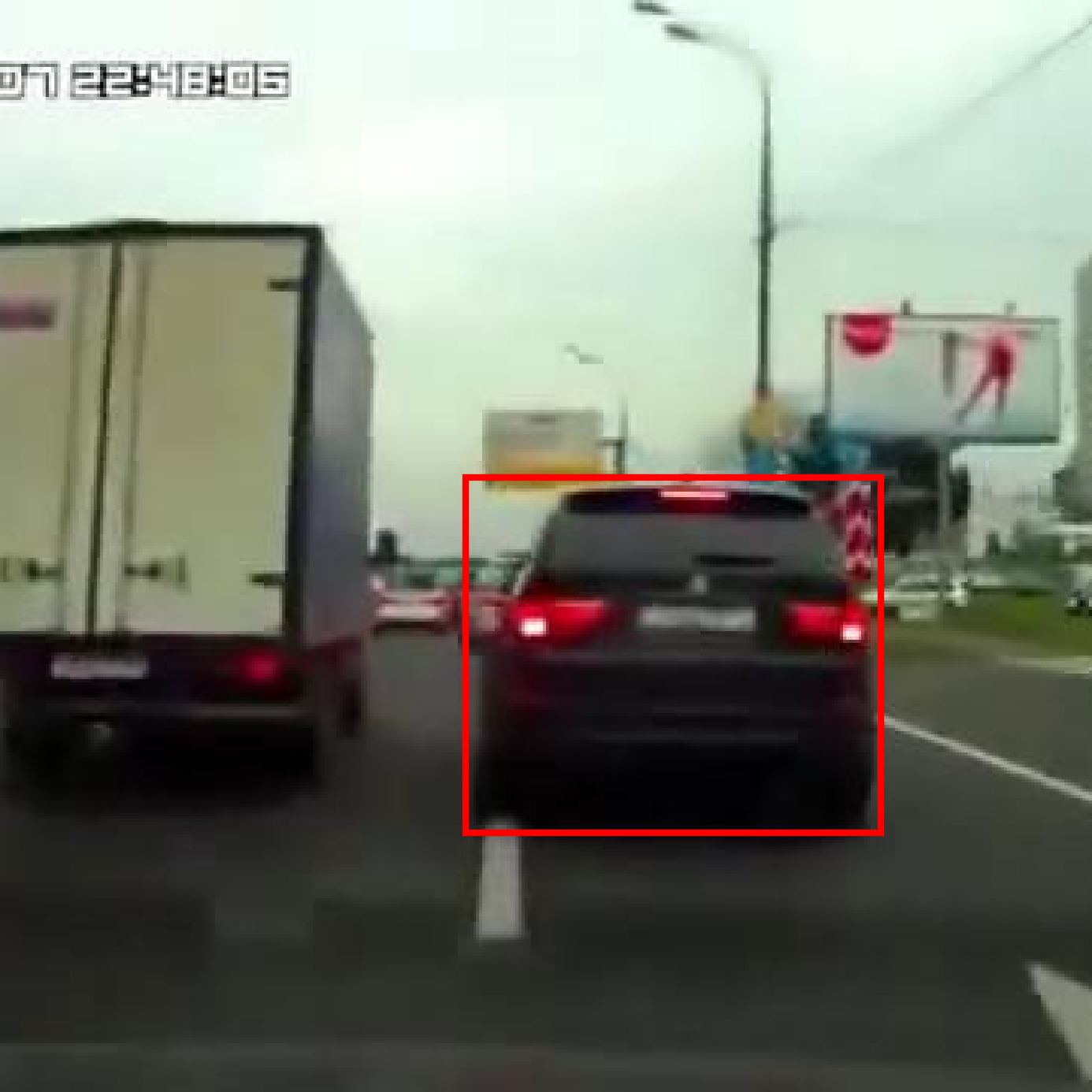}}};
        \node[inner sep=0pt, label=above:$t_1$] (frame2) at (4.2,0) {\fbox{\includegraphics[scale=0.23]{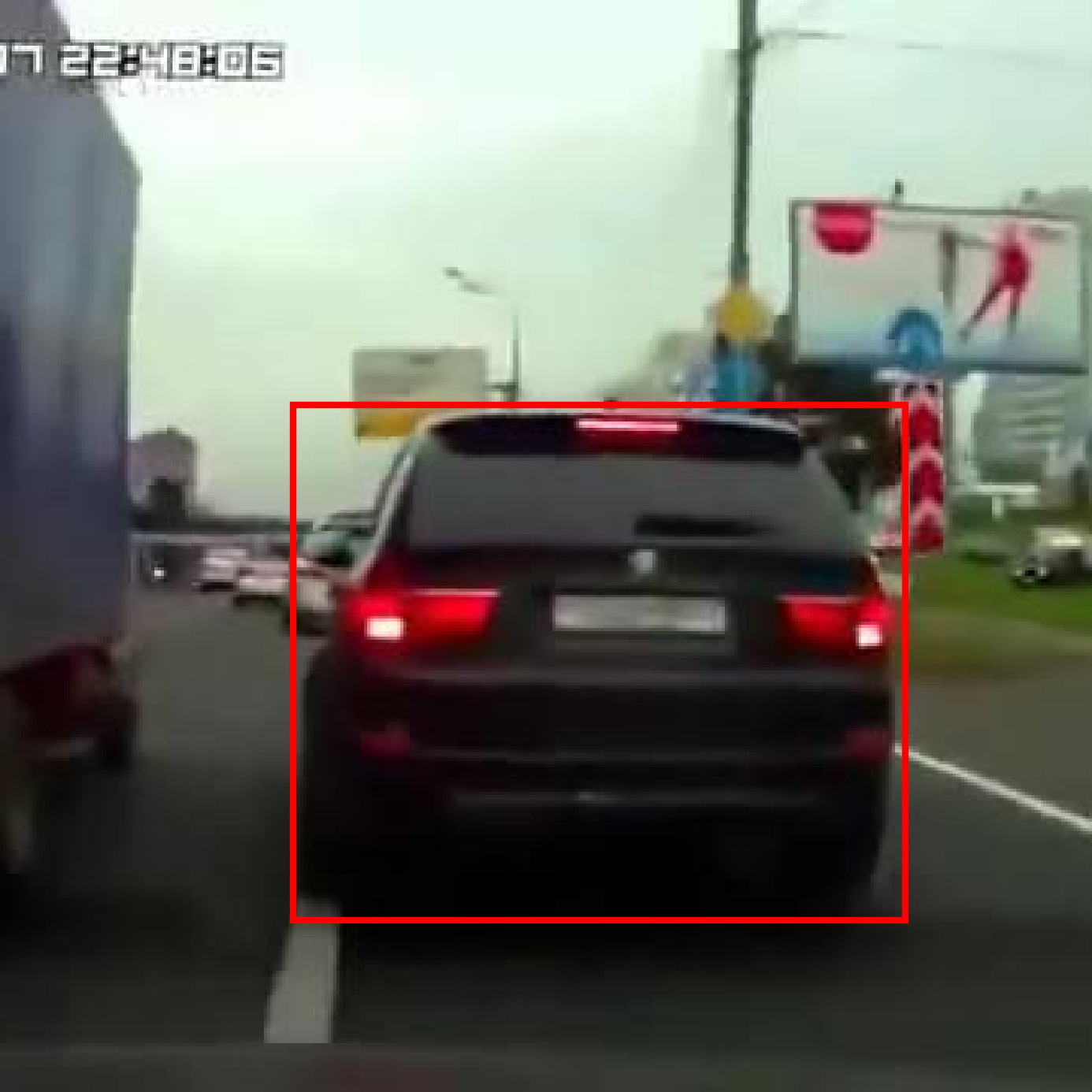}}};
        \node[inner sep=0pt, label=above:$t_2$] (frame3) at (8.4,0) {\fbox{\includegraphics[scale=0.23]{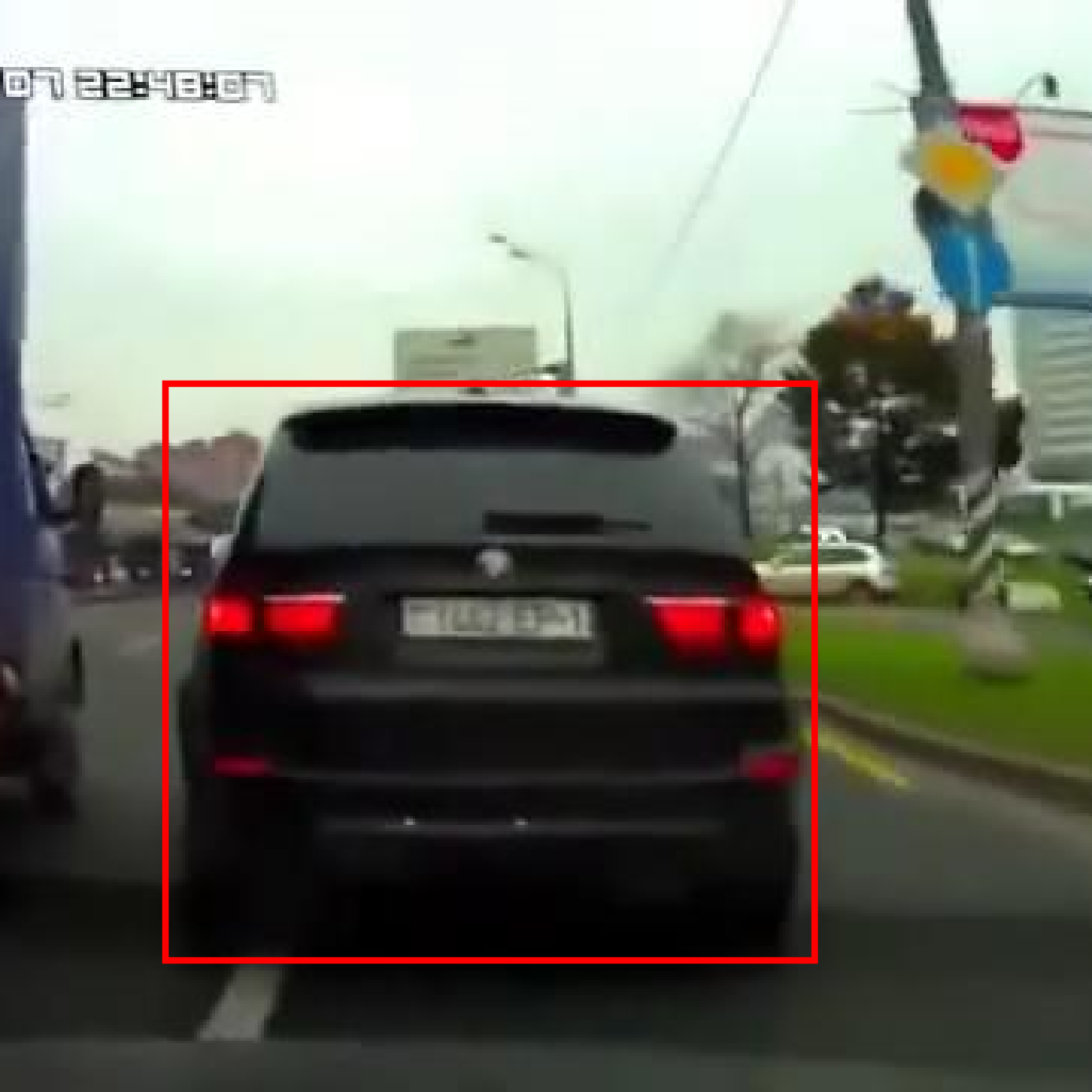}}};
        \node[inner sep=0pt, label=above:$t_3$] (frame4) at (12.6,0) {\fbox{\includegraphics[scale=0.23]{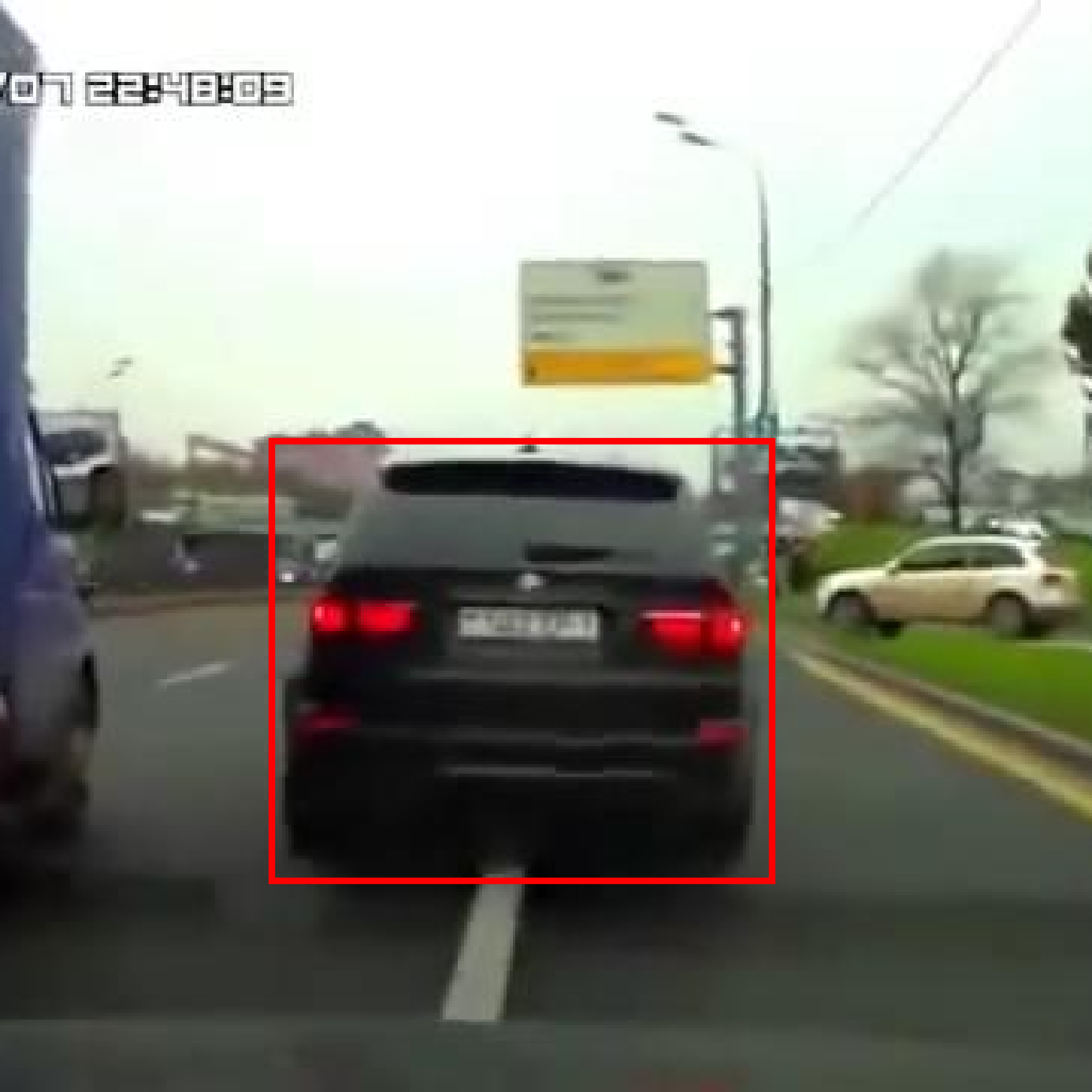}}};

        \begin{scope}[on background layer]
            \draw[rounded corners, fill=red!10] (-1.5, 1.5) rectangle ++(15.8, -3);
        \end{scope}

        \node[inner sep=0pt] (drise) at (-2.3, -3.2) {\drise};
        \node[inner sep=0pt] (dr1) at (0.,-3.2) {\includegraphics[scale=0.27]{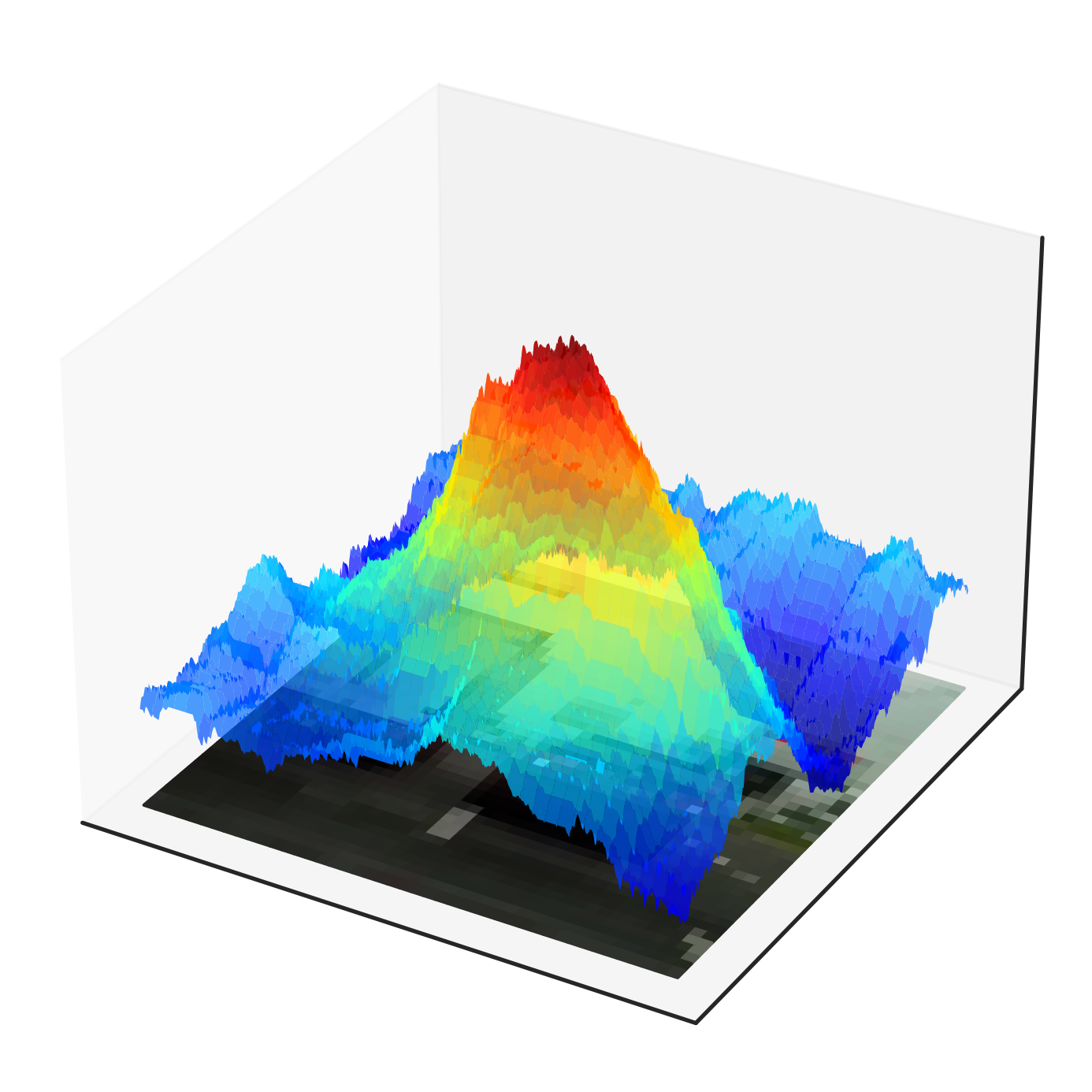}};
        \node[inner sep=0pt] (dr2) at (4.2,-3.2) {\includegraphics[scale=0.27]{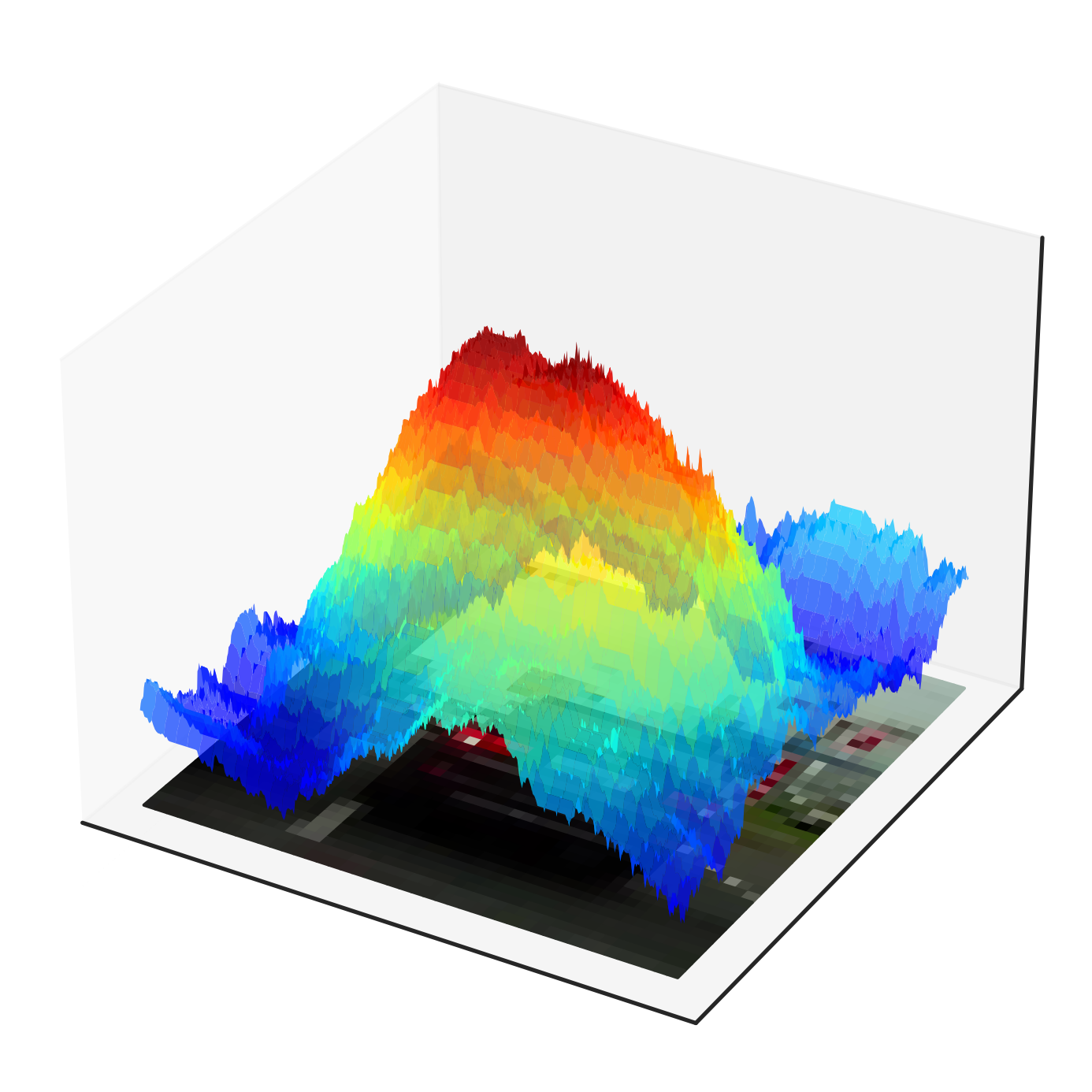}};
        \node[inner sep=0pt] (dr3) at (8.4,-3.2) {\includegraphics[scale=0.27]{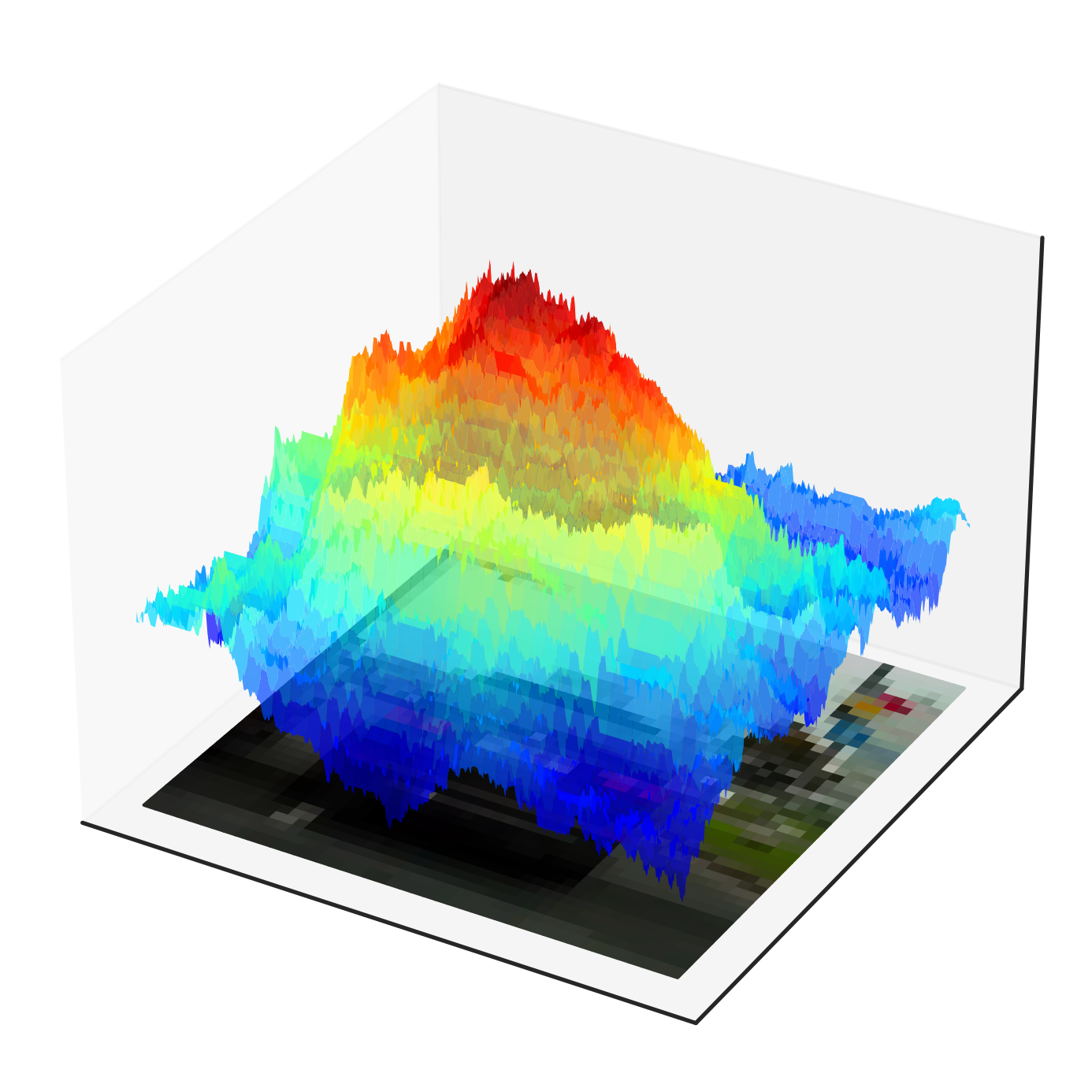}};
        \node[inner sep=0pt] (dr4) at (12.6,-3.2) {\includegraphics[scale=0.27]{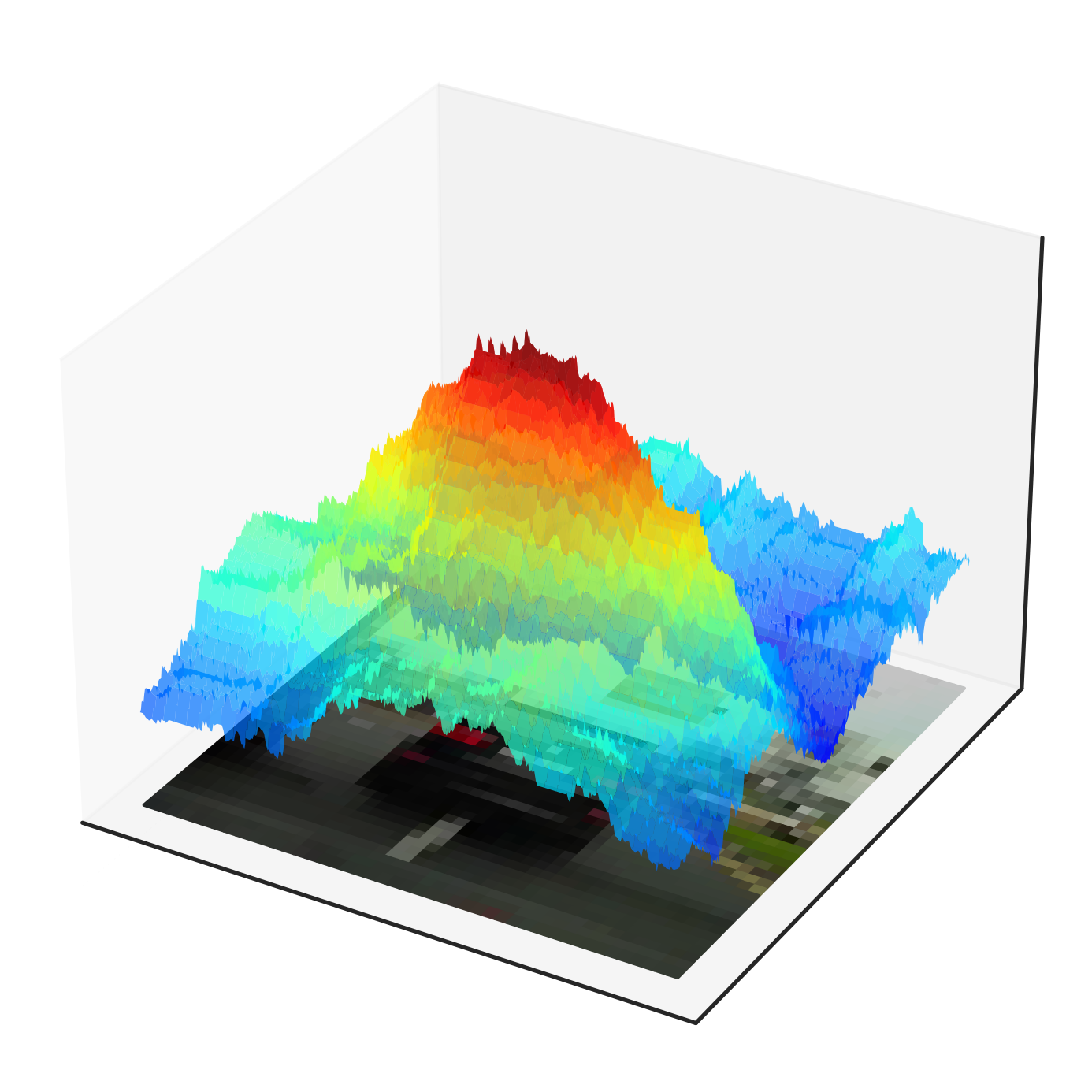}};

        \begin{scope}[on background layer]
            \draw[rounded corners, fill=green!05] (-1.5, -1.75) rectangle ++(15.8, -3);
        \end{scope}

        \node[inner sep=0pt] (drise) at (-2.3, -8.1) {\incx};
        \node[inner sep=0pt] (incx1) at (0,-6.5) {\includegraphics[scale=0.27]{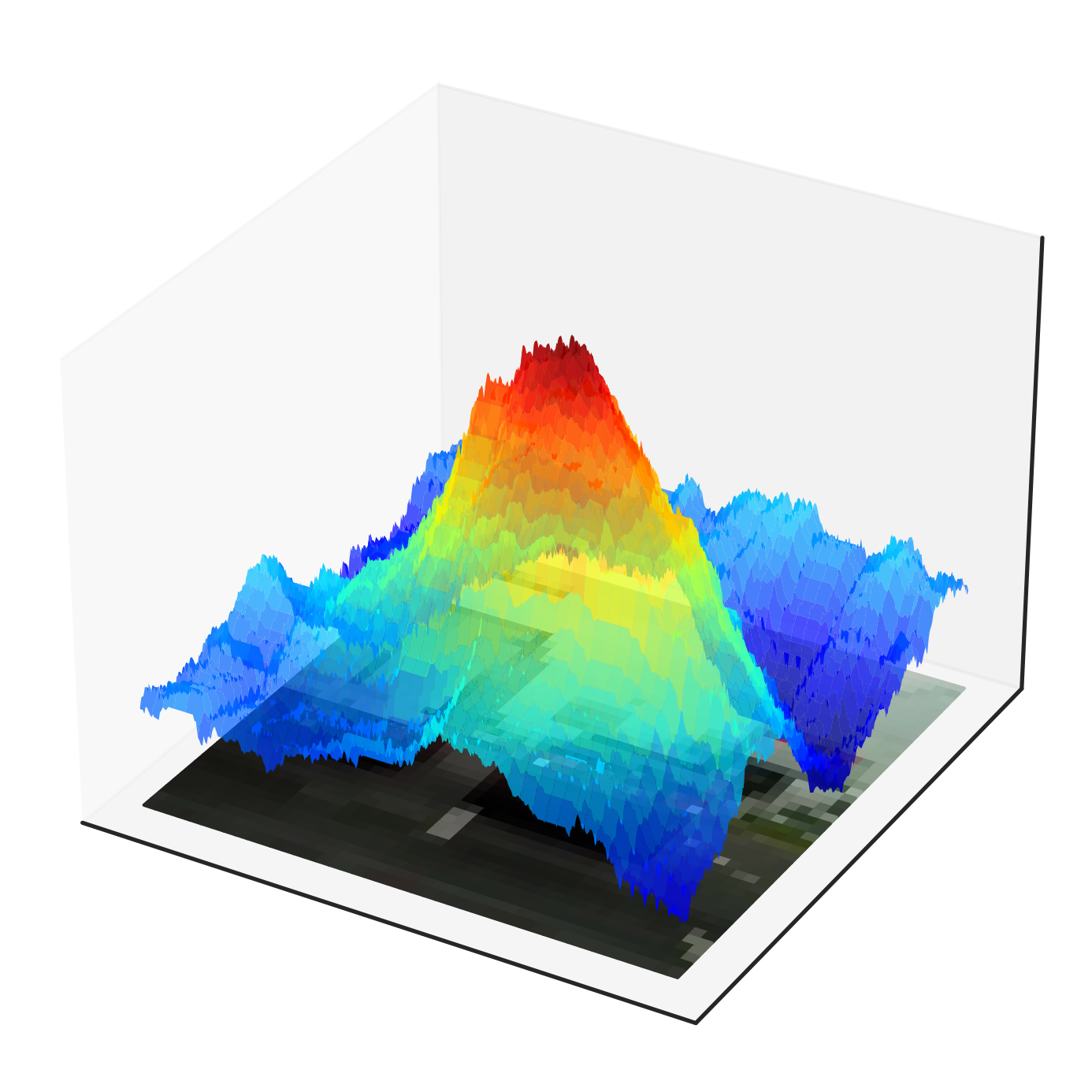}};
        \node[inner sep=0pt] (incx2) at (4.2,-6.5) {\includegraphics[scale=0.27]{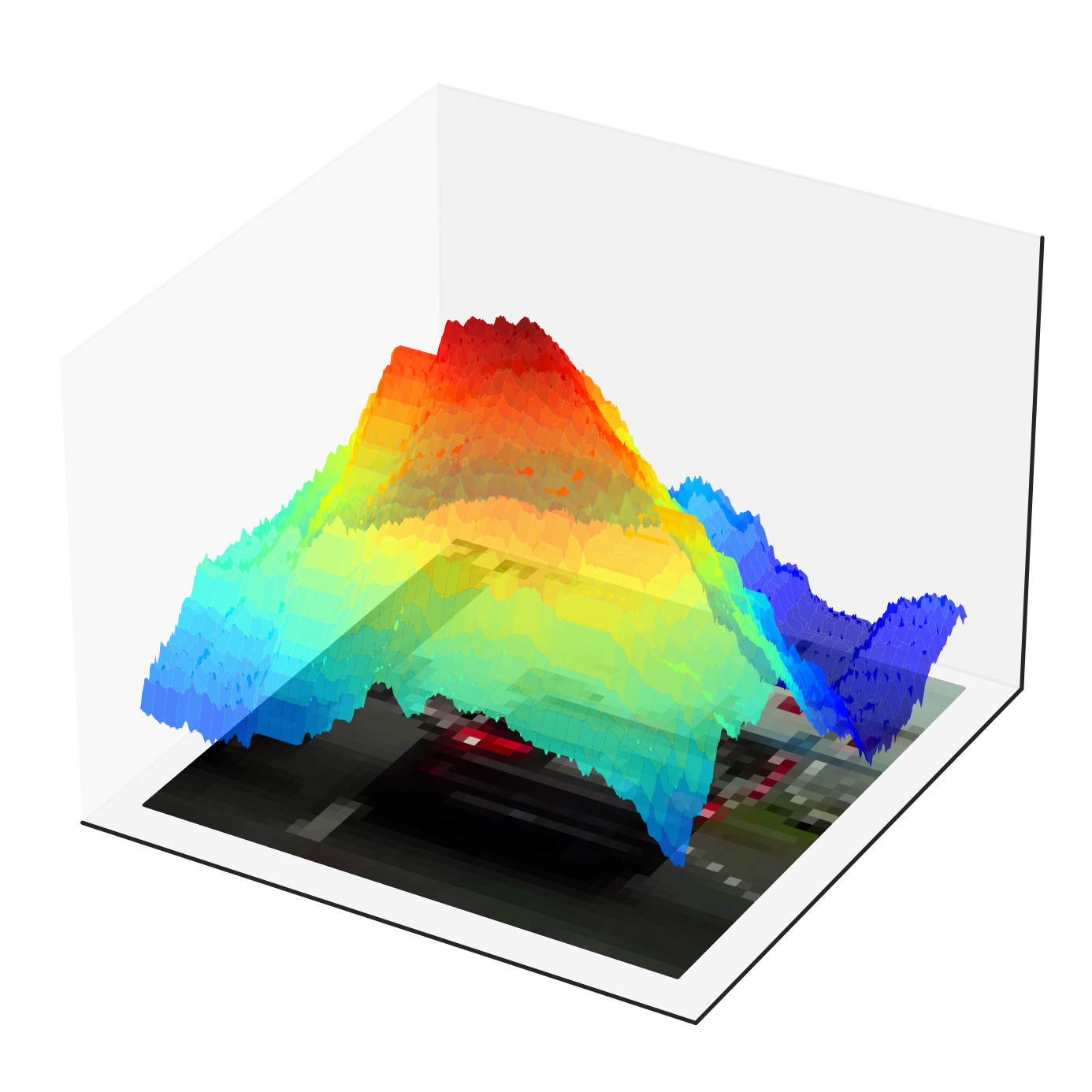}};
        \node[inner sep=0pt] (incx3) at (8.4,-6.5) {\includegraphics[scale=0.27]{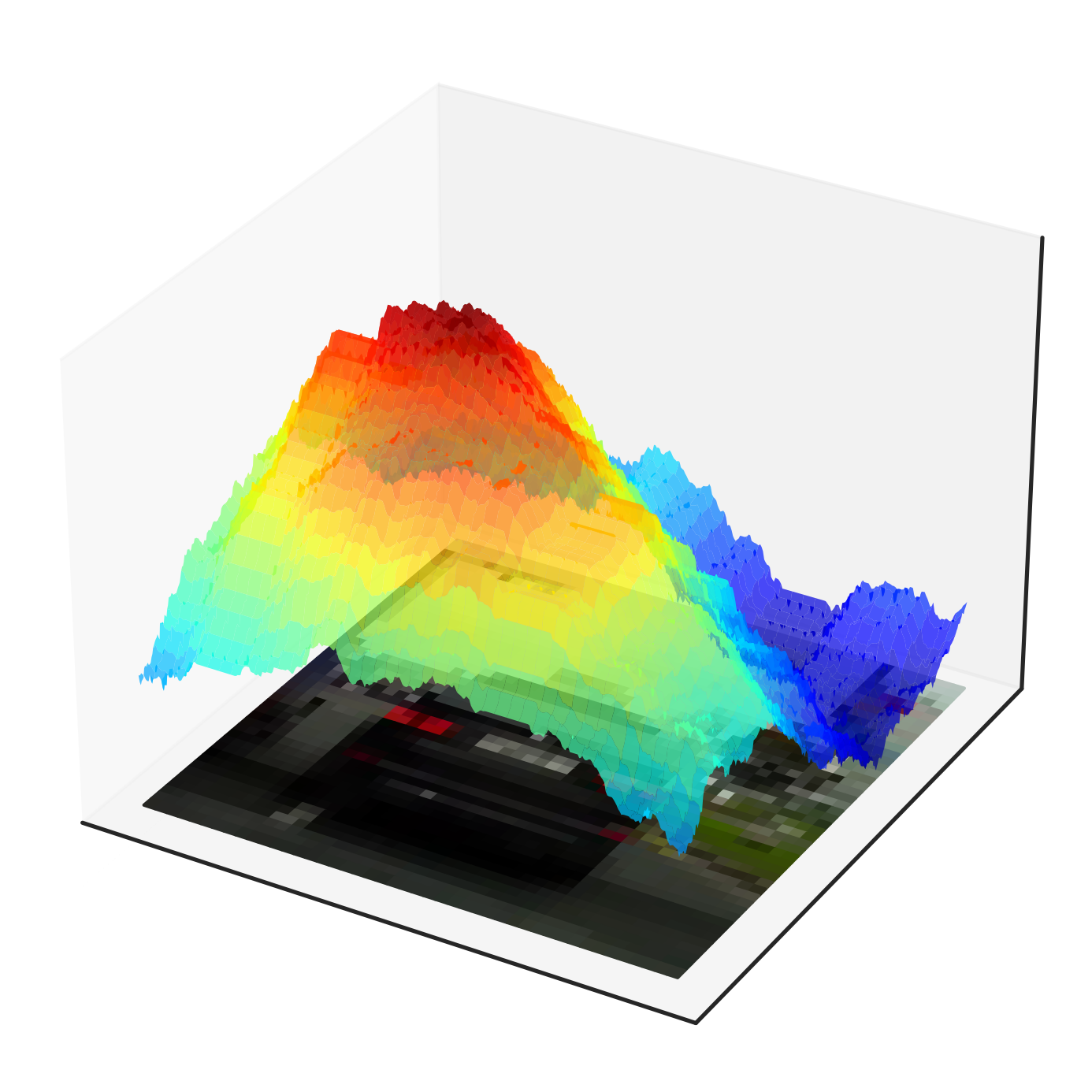}};
        \node[inner sep=0pt] (incx4) at (12.6,-6.5) {\includegraphics[scale=0.27]{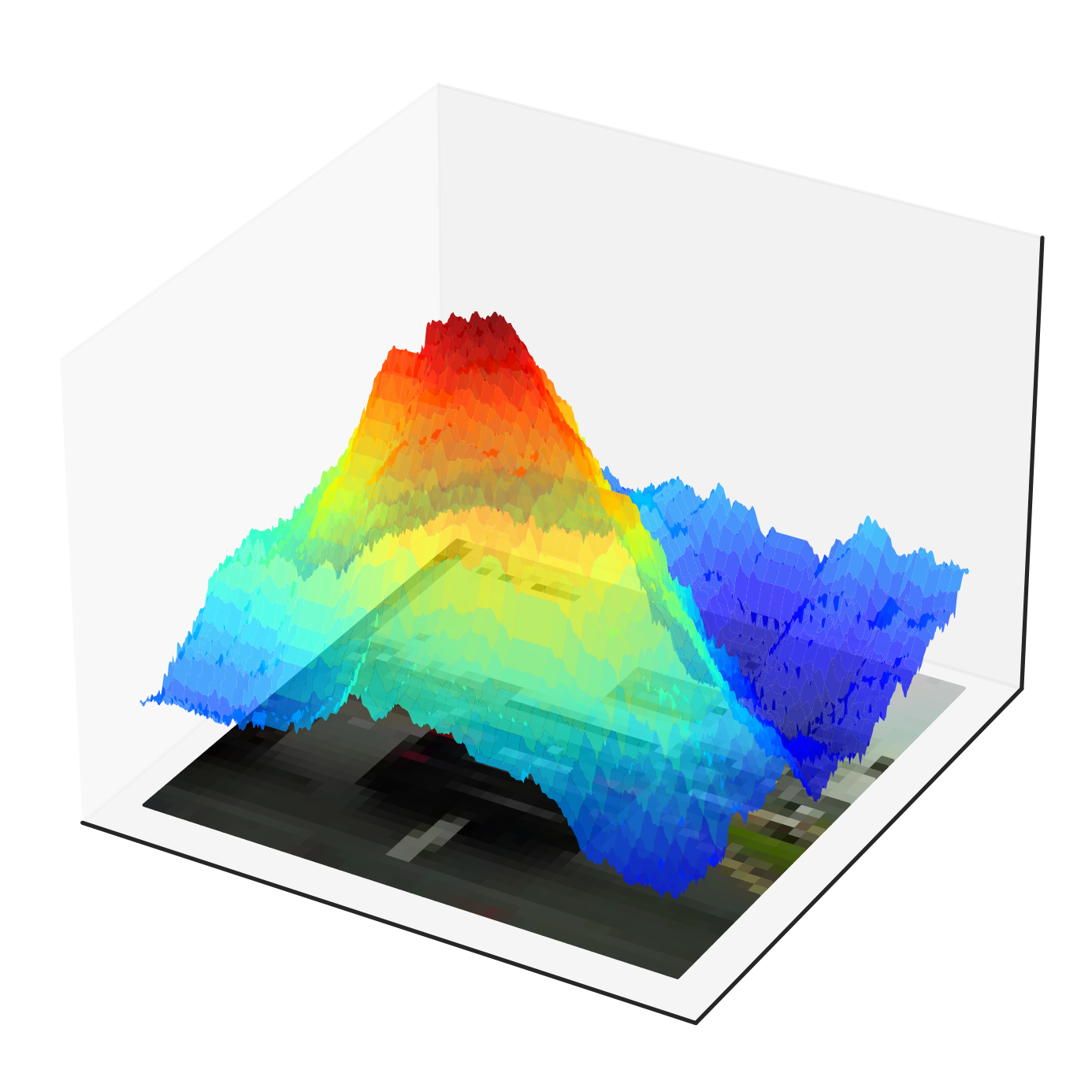}};

        \draw (incx1) edge["$V_{t0}\! \rightarrow \! V_{t1}$"] (incx2);
        \draw (incx2) edge["$V_{t1}\! \rightarrow \! V_{t2}$"] (incx3);
        \draw (incx3) edge["$V_{t2}\! \rightarrow \! V_{t3}$"] (incx4);

         \begin{scope}[on background layer]
            \draw[rounded corners, fill=blue!05] (-1.5, -5) rectangle ++(15.8, -3);
        \end{scope}

        \node[inner sep=0pt] (exp1) at (0,-9.8) {\fbox{\includegraphics[scale=0.21]{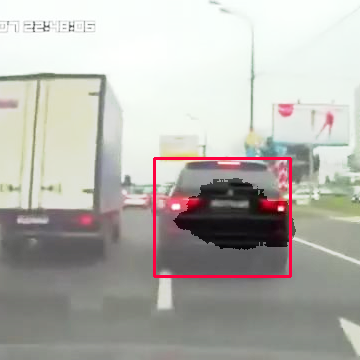}}};
        \node[inner sep=0pt] (exp2) at (4.2,-9.8) {\fbox{\includegraphics[scale=0.21]{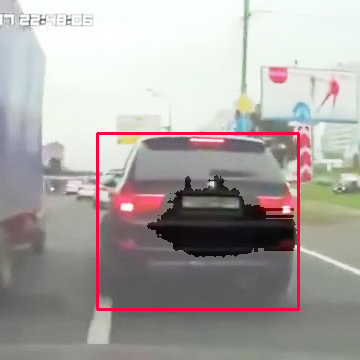}}};
        \node[inner sep=0pt] (exp3) at (8.4,-9.8) {\fbox{\includegraphics[scale=0.21]{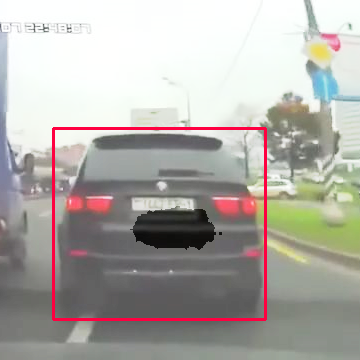}}};
        \node[inner sep=0pt] (exp4) at (12.6,-9.8) {\fbox{\includegraphics[scale=0.21]{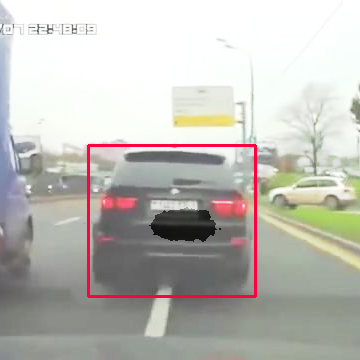}}};
        
         \begin{scope}[on background layer]
            \draw[rounded corners, fill=blue!05] (-1.5, -8.3) rectangle ++(15.8, -3);
        \end{scope}
        
    \end{tikzpicture}
    \caption{Detection of an object `car' (Frames) at four different time stamps, with \drise saliency landscapes and the approximate landscapes and explanations generated by \incx, 
    showing the similarity between the fully computed \drise saliency maps and the estimated \incx ones. $V_t$ represents the object at time $t$.}
    \label{fig:accept}
\end{figure*}
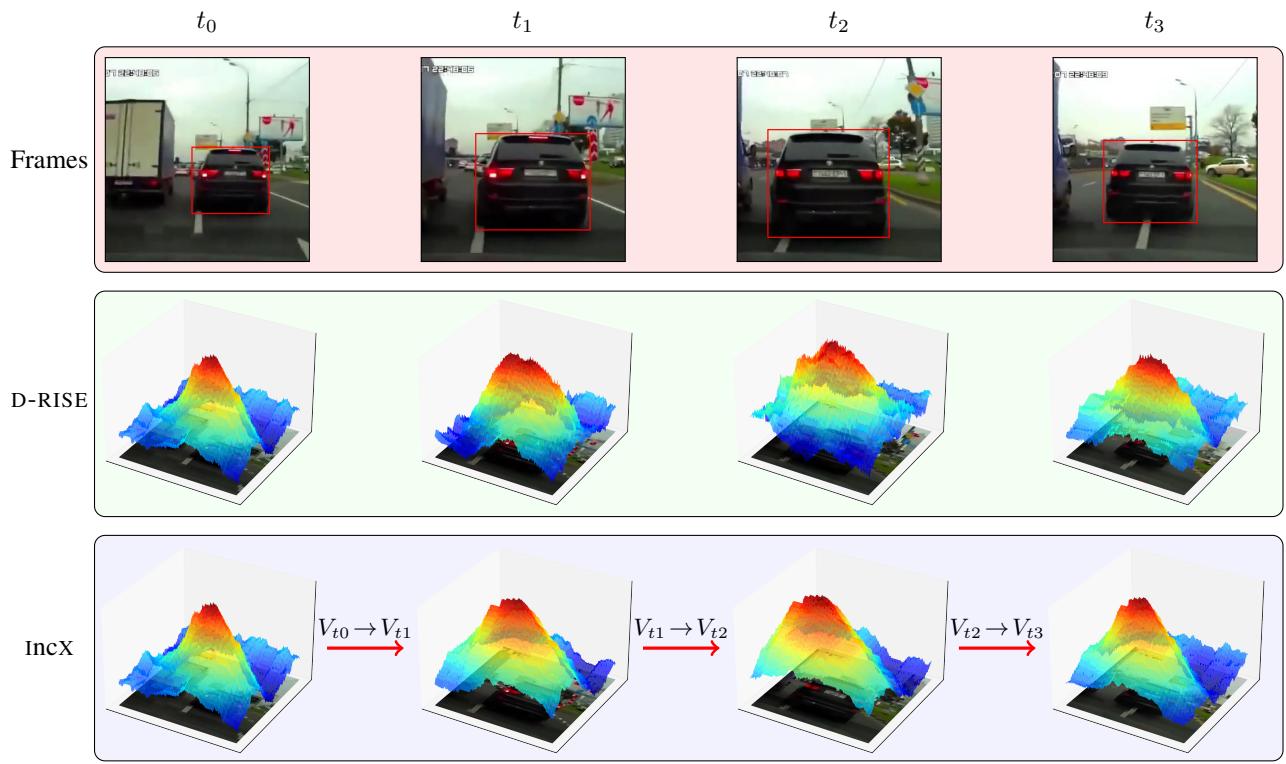






AI models are now a primary building block of most computer vision systems. However, the inherent opacity of some of these models (e.g., neural networks) and their sharp increase in size and complexity, slow their adoption, especially in safety-critical applications, due to the need to understand the reasoning of the model~\cite{AnanthaswamyAnil2023InBetter}.

Explainable AI (\xai) is used to provide insights into the decision-making process of machine learning models. Post hoc \xai can be roughly divided into ``white-box'' and ``black-box'' methods. White-box methods have access to the internals of the model, and are hence tailored to a particular model architecture. Black-box methods can only access model outputs and are agnostic to the internal implementation of the model.
In this paper we focus on black-box explainability, due to its portability across models.

Explainable AI is particularly relevant in the context of autonomous driving, as it is instrumental in assessing safety of autonomous vehicles. 
Regulatory bodies can greatly benefit from XAI by improving model explainability, which aids in product liability litigation and compliance assessments~\cite{Deeks2019}. Furthermore, industry standards for road vehicle safety, such as ISO21448~\cite{ISO21448:2022}, emphasize the need for traceability of performance insufficiencies and their trigger conditions. In this regard, explainable AI has been proposed as a means of enhancing implementation transparency, contributing to the development of inherently safe AI systems in autonomous driving (AD)~\cite{mohseni2019practicalsolutionsmachinelearning}.

Object detection is crucial in AD for various tasks, including traffic sign and light recognition~\cite{Jinkyu2018}. 
Visual explanations, such as saliency maps, are essential for enhancing transparency and trust in autonomous driving systems~\cite{atakishiyev2024explainable}. They aid in generating textual observations and improving system interpretability \cite{Kim2021}.
Explaining object detectors, however, is challenging. On the one hand, most AD approaches rely on 
white-box algorithms~\cite{Nowak2019,Bojarski2018}, which are primarily based on gradient propagation, restricting their applicability to specific architectures. 
On the other hand, while black-box methods are model-agnostic, their high computational cost makes real-time 
explanations infeasible~\cite{MORADI2024109183}.

To enhance autonomous driving models' explainability, we introduce \emph{\underline{Inc}remental e\underline{X}planations}, or \incx\textemdash a 
black-box \xai algorithm and tool for explaining object detectors in real-time. \incx avoids multiple calls to the model
and thus has a negligible overhead over the object classification running time, allowing near real-time explanation processing for video data.
\incx tracks saliency maps from one frame to the next, applying linear transformations to the saliency map generated in the first frame,
and uses these maps to compute sufficient explanations\textemdash subsets of pixels of the image sufficient for the original classification.
Unlike \fastshap~\cite{fastshap21}, it does not rely on a separate machine learning model that would necessitate, in turn, its own explanations.

We show that \incx computes explanations that are comparable in quality to \drise~\citep{drise}, 
but outperforms \drise in speed by two orders of magnitude. To the best of our knowledge, 
this is the first black-box real-time \xai tool for object detectors. 

The code, the benchmark sets, full proofs, and the full tables of results are submitted as a part of the supplementary material. Instructions for running \incx locally in a docker container and seeing the explanations in real time are also provided.

\section{Related Work}\label{sec:relwork}

We survey the related work in object detection, tracking, and explainability. Since this work focuses on black-box explainability, we do not
include white-box \xai methods, such as GradCAM~\cite{CAM} or DexT~\cite{PPAV23}. 

\paragraph*{Explainability for image classifiers} There is a large body of work on explainability for image
classifiers. 
Common black-box tools are \lime~\cite{Ribeiro2016WhyClassifier}, 
\shap~\cite{Lundberg2017APredictions}, \rise~\cite{petsiuk2018rise}, and \rex~\cite{CKKS24}. 
These tools and algorithms define and output explanations in different formats: functions, models and landscapes. There is a straightforward conversion of different formats of explanations to \emph{saliency maps}, which rank input pixels based on their contribution to the model classification. 
Black-box tools generally rely on multiple calls to the model, computing explanations from these results, and hence incurring a significant overhead on top of the image classifier.

\paragraph*{Object detection in AD} The object detection task is finding objects and classifying 
them under low latency. This capability is fundamental in autonomous driving, as it enables the vehicle to recognize other automobiles and traffic signs
in real time.

\yolo~\cite{yolo}, a collection of one-stage algorithms, has enjoyed a considerable popularity due to its balance between efficiency and accuracy, leading to adoption in autonomous driving research~\cite{Alaahdal20242792}. It divides the image into a grid of cells and assigns a set of class probabilities, a bounding box, and a confidence value. Then it performs non-maximum suppression to remove overlapping bounding boxes. The 
latest version, $10$th as of time of writing, addresses
many of the shortcomings of the original \yolo~\cite{Wang2024YOLOv10:Detection}.

In contrast, a two-stage process uses a region proposal network (RPN) to identify regions within the image that contain an object. 
Then it takes those regions and warps them into a classifier network to find the probabilities of the objects inside the image. Examples of such architectures include \fastrcnn~\cite{Ren2015FasterNetworks}, \rcnn~\cite{Girshick2014RichSegmentation} and \frcnn~\cite{Girshick2015FastR-CNN}. This approach has also been extensively used in AD research~\cite{Liang_2024_CVPR}.

Transformer-based detectors have been popular since the introduction of the transformer architecture, which revolutionized the Natural Language Processing (NLP) domain, as it introduced the self-attention mechanism~\cite{Vaswani2017AttentionNeed}. The Real-Time Detection Transformer (\rtdetr) leverages the same concept for object detection 
tasks~\cite{zhao2024detrs}.

\paragraph*{Explainability for object detectors}
Several XAI algorithms can be modified to provide explainability for object detectors. 
There are only a few black-box algorithms that generate saliency maps without relying on the model architecture. 
\drise~\cite{drise} is a modification of \rise for black-box object detectors, using the basic underlying idea of \rise to 
generate a set of random masks. \drise also introduces the concept of a detection vector, composed of localization 
information (bounding boxes), object score, and classification information. It then uses this vector to compute a weight for 
each mask and construct a saliency map given by the weighted sum of masks.
In contrast, \fsod~\cite{Kuroki2024FastDetection} adapts the loss function of \fastshap to train an explainer model with an architecture inspired by UNet to account for classification and location of the detected object~\cite{Jethani2022FastSHAP:Estimation}. Although this approach exhibits a considerably faster performance than the previously mentioned ones, its major drawback is that it lacks explainability for the explainer model itself. 

\paragraph*{Multiple Object tracking (MOT)}
Detection-based tracking (DBT) is the dominant paradigm in multiple object tracking tasks. These methods divide the tracking sub-problem into three main tasks: object detection, re-identification (ReID) and trajectory tracking~\cite{Du2024ExploringDirections}. 
%
A classic example of a Multi-object tracker within this paradigm is SORT~\cite{Bewley2016SimpleTracking}, which is the algorithm we use in this paper. SORT uses a Kalman filter to perform estimations of previously detected objects. Subsequently, it associates the identified objects via the Intersection over Union (IoU) of the bounding box of the detected object and the estimated position and scale of the bounding box. It uses the Hungarian algorithm to optimally solve this assignment. 
There is a number of algorithms implementing improvements to SORT, such as DeepSORT~\cite{Wojke2017SimpleMetric}, StrongSORT~\cite{Du2023StrongSORT:Again},
and ByteTrack~\cite{Zhang2022ByteTrack:Box}, but these rely on an additional deep learning model, detracting from transparency.


\paragraph*{Explainable AI in Autonomous Driving} Explainable AI for safe autonomous driving can be categorized into several approaches~\cite{Kuznietsov2024}. Interpretable by design models are inherently transparent, such as traffic sign detectors built with Inductive Logic Programming~\cite{Chaghazardi2023}. Interpretable surrogate models approximate black-box behavior with explainable alternatives like LIME~\cite{Ribeiro2016WhyClassifier}. Interpretable monitoring employs saliency maps to track model behavior and detect anomalies~\cite{Hacker2023}. Interpretable validation ensures reliability through offline interpretable methods. Together, these techniques strengthen AI transparency and safety in autonomous driving.

\section{Incremental Explanations}\label{sec:incx}

In this section we present our contribution -- the \incx algorithm.
Full proofs of all theorems and lemmas are provided in the supplementary material. 

\begin{figure*}[t]
    \centering
    \includegraphics[scale=0.525]{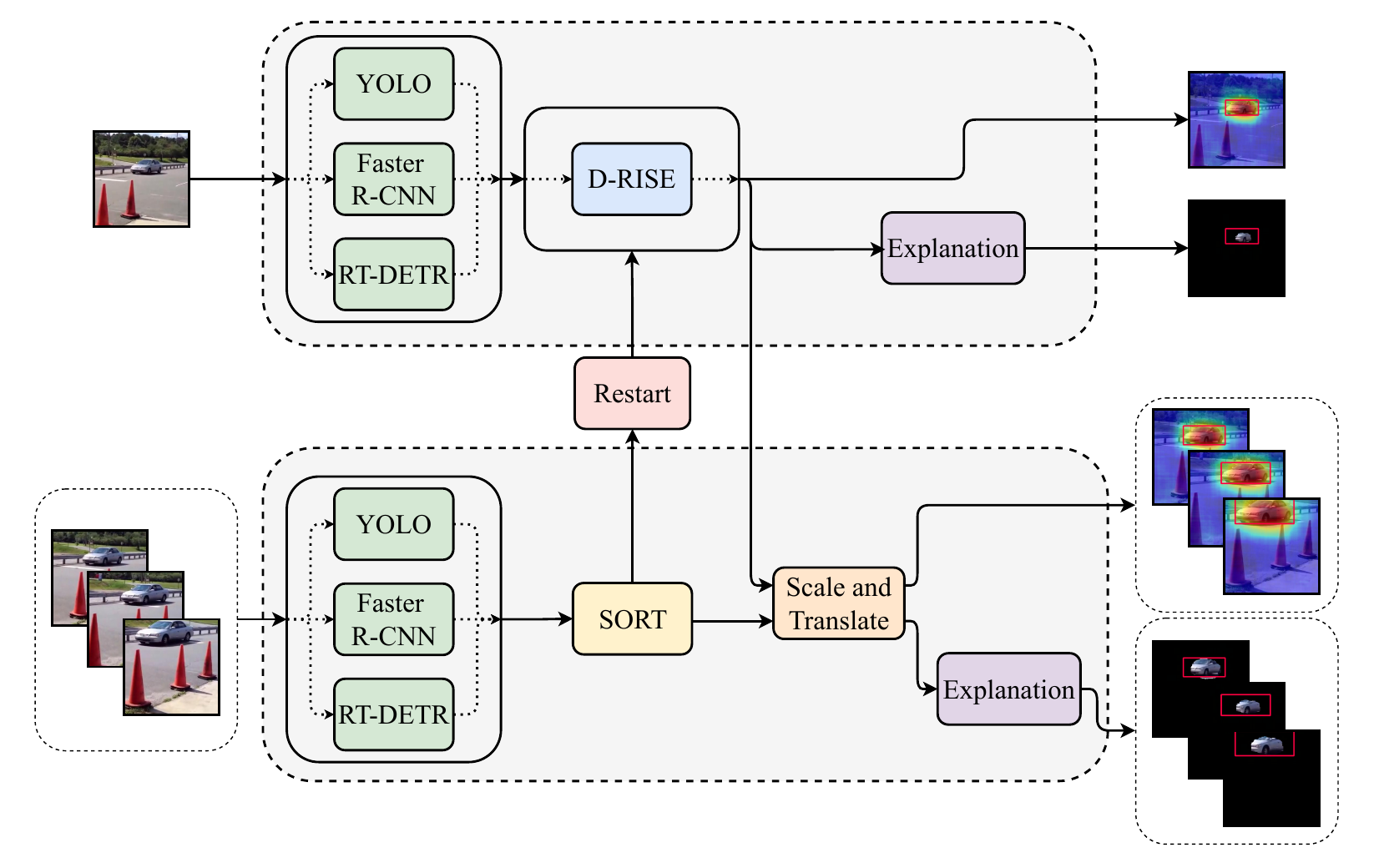}
    \caption{Block diagram of \incx components: First frame process (Top) and subsequent frames (Bottom)}
    \label{fig:increx}
\end{figure*}

\subsection{Theoretical Foundations}
The intuition behind our algorithm is based on the following lemma. 
\begin{lemma}
For a fixed observer, movement of an object in $3D$ space without rotation or deformation can only result in a combination of
scaling and linear translation when projected on a given vertical plane.
\end{lemma}
\begin{proofsketch}
The lemma follows from the observation that the vector of movement of an observed object is a sum of its projections to the vertical plane and
in the direction to or away from the observer. The vector to or away from the observer determines the scaling, while the projection to
the vertical plane determines the linear translation. 
\end{proofsketch}

Given a transformation of the observed object as above and its initial saliency map, we prove that the saliency map of the transformed
object can be computed from the original saliency map using a similar transformation.

The main result is stated in \Cref{thm:translation}. The formal proof is based on results from probability theory and is quite involved (see the supplementary material). Before presenting the theorem, we first introduce additional context and formalization to ensure a clearer understanding.

Intuitively, the proof is based on converting the saliency map to a \emph{probability mass function} (\emph{pmf}) and then applying the mathematics of pmf transformation to estimate a pmf for a new, updated, saliency map. 

First, the initial saliency map is converted into a pmf by normalizing the values of the saliency map.
Then we use ~\Cref{thm:prob_conv}~\cite{Hogg2019IntroductionStatistics} to define the transformations of the pmf. 
This allows us to calculate the new saliency pmf after undergoing linear transformation.
Technically, our pmf is a joint probability mass function $P(X, Y)$.
We restate the core transformation result over a single variable from~\citet{Hogg2019IntroductionStatistics} here for completeness. 
\begin{lemma}[\cite{Hogg2019IntroductionStatistics}]
\label{thm:prob_conv}
Assume $X$ to be a discrete random variable with space $D_X$ and $g$ a one-to-one transformation function. Then the space of $Y$ is 
$D_Y = \{g(x) : x \in D_X\}$, and the pmf of $Y$ is
\[
p_Y(y) = P[g(X) = y] = P[X = g^{-1}(y)] = p_X(g^{-1}(y)).
\]

\end{lemma}

\Cref{thm:prob_conv} can be extended to joint pmfs over multiple variables~\cite[p.100]{Hogg2019IntroductionStatistics} (see supplementary material).



Having established the intuition behind the transformation of the saliency pmf, we address the considerations of our approach concerning transformations involving three-dimensional objects. Specifically, we first describe how such objects get projected onto the image plane. By~\citet{Forsyth2002ComputerVision}, the mapping of points from three-dimensional world coordinates to two-dimensional screen coordinates can be achieved while using a common frame of reference for the world and the camera. Lemma \ref{lem:projection} formalizes this projection process.
\begin{lemma}
\label{lem:projection}
Let \(\alpha, \beta \in \mathbb{R}\) denote the intrinsic magnification factors of the camera. The transformation from 3D world coordinates \((x, y, z)\) to 2D screen coordinates \((u, v)\) is given by:

\[ \begin{bmatrix}
u \\ v
\end{bmatrix} = \frac{1}{z} \begin{bmatrix}
\alpha & 0 & O_x \\
0 & \beta & O_y
\end{bmatrix} \begin{bmatrix}
x \\
y \\
z
\end{bmatrix}= \frac{1}{z} \mathcal{K} \begin{bmatrix}
x \\
y \\
z
\end{bmatrix} \]

where \(\mathcal{K}\) represents the intrinsic parameters of the camera, and \((O_x, O_y)\) represents the coordinates at the upper-left corner of the image.
\end{lemma}

We formalize the intuition of a \emph{center} of a bounding box for the projected object in~\Cref{def:center}. 
This concept is leveraged in the scaling and translation operations described in \Cref{def:sca_trans}, which are fundamental for \incx.
\begin{definition}[Center Function]
\label{def:center}
The center function $\mathbf{C}: 2^{\mathbb{R}^n} \to \mathbb{R}^n$ maps a set $S \subset \mathbb{R}^n$ of points to the center point of the bounding box that encloses these points. For a set $S$ consisting of points $\mathbf{p} = (p_{1}, p_{2}, \ldots, p_{n}) \in \mathbb{R}^n$ the function is defined as follows:

\[
\mathbf{C}(S) = \begin{bmatrix}
\frac{\max(p_{1}) + \min(p_{1})}{2} \\
\frac{\max(p_{2}) + \min(p_{2})}{2} \\
\vdots \\
\frac{\max(p_{n}) + \min(p_{n})}{2}
\end{bmatrix}
\]

where $max(p_j)$ and $min(p_j)$ denote the maximum and minimum values, respectively, of the $j$-th coordinate among all points in $S$.
\end{definition}
\noindent



\begin{definition}[Scaling and Translation]
\label{def:sca_trans}
Let \( A = \{(u, v) \in \mathbb{R}^2 \} \) represent a set of points defining an area in the image space.

\noindent
\textbf{Scaling} is a linear transformation $S: \mathbb{R}^2 \to \mathbb{R}^2$, defined as \( S(\mathbf{p}) = \gamma(\mathbf{p} - \mathbf{C}(A)) \), where \( \mathbf{p} \in A \) and \( \gamma \in \mathbb{R}^{2 \times 2} \) is a diagonal matrix. 

\noindent
\textbf{Translation} is a linear transformation $T: \mathbb{R}^2 \to \mathbb{R}^2$, defined as $T(\mathbf{p}) = \mathbf{p} + \mu$, where $\mathbf{p}\in A$. 
\end{definition}

The scaling transformation \( S \) modifies a point \( \mathbf{p} \) 
by a factor of \( \gamma \), after first shifting it relative to the center of the area of the object in the image. Specifically, applying \( S \) to all points in \( A \) is equivalent to scaling the entire image while moving the center of mass \( \mathbf{C}(A) \) to the origin \( (0,0) \).
The translation function \( T \) \emph{shifts} the two-dimensional point \( \mathbf{p} \) by a vector \( \mu \in \mathbb{R}^2 \). Similarly, applying this transformation to all points in an image results in translating the entire image by \( \mu \).

We first provide the intuition behind our theorem and then its formal statement in~\Cref{thm:translation}.
Our approach maps a given pmf to a $2D$-projection of a three-dimensional object $\delta_t$ (see 
Assumption~\ref{assum:expected_value}). Furthermore, as long as the object being tracked does not leave the frame, 
rotate, or is occluded by another object, the future positions of the object are determined solely 
by translation and scaling transformations in the three-dimensional space 
(see Assumption~\ref{assum:three_dimensional}).
Together, these assumptions lead to the statement that future saliency maps can be computed by 
applying a scaling and translation linear transformation to the current saliency map.

We use the following notation. 
Let \( \mathbf{Q}_t = (Q_{1,t}, Q_{2,t}) \) be a vector in the image space at time $t$, 
where $Q_{1,t}$ and $Q_{2,t}$ represent random variables along the $u$-axis and the $v$-axis, respectively. 
The pmf of $\mathbf{Q}_t$ is \( p_{\mathbf{Q}_t}(\mathbf{q}_{t}) \), which is derived from the normalized saliency map.




We assume that the expected value of the saliency pmf, if it exists, is equal to the center of the bounding box 
$\mathbf{C}(\mathcal{K}V_t)$ of the object $V_t$ as projected into the image space using the matrix $\mathcal{K}$. 
$\delta_t$ represents the distance between this projected center and the expected value of $\mathbf{Q}_t$. 
The main reason behind this assumption is the homogeneity of our pmf, implying that any scaling or translation 
of the expected value results in a corresponding scale or shift of the entire pmf, preserving the distribution's
proportions. A direct implication of this assumption is that if we move and scale the center of the bounding box, 
then the saliency map would have a similar shift and scaling.

\begin{assumption}[Expected Value and Bounding Box Center Relation] \label{assum:expected_value}
\[
\mathbb{E}[\mathbf{Q}_t] = \mathbf{C}(\mathcal{K}V_t) + \delta_t
\]
\end{assumption}

\begin{assumption}[Constrained Affine Transformation] \label{assum:three_dimensional}
Any point of an object at time $t+1$ is related to the same object at time $t$ by an affine transformation given by:

\[
\begin{bmatrix}
    x_{t+1} \\
    y_{t+1} \\
    z_{t+1} \\
    1
\end{bmatrix}
=
\begin{bmatrix}
    m_x & 0 & 0 & 0 \\
    0 & m_y & 0 & 0 \\
    0 & 0 & m_z & 0 \\
    0 & 0 & 0 & 1
\end{bmatrix}
\begin{bmatrix}
    \mathbf{I} & \mathbf{t} \\
    \mathbf{0}^T & 1
\end{bmatrix}
\begin{bmatrix}
    x_t \\
    y_t \\
    z_t \\
    1
\end{bmatrix}
\]
where $\mathbf{t} \in \mathbb{R}^3$ is a translation vector, $\mathbf{I} \in \mathbb{R}^{3 \times 3}$ is the identity matrix, and $m_x$, $m_y$, and $m_z \in \mathbb{R}$ are scaling factors.
\end{assumption}

Finally, let $S$ be the scaling function, $T$  the translation function, $\mathbf{q}_{t+1}$ a coordinate vector 
in the image space at time $t+1$, and $\mathbf{q}_{t}$ a coordinate vector at time $t$. 
The following theorem describes the relation between the pmf at time $t$ and at time $t+1$ and is instrumental 
to our algorithm.
\begin{theorem}\label{thm:translation}
If Assumptions~\ref{assum:expected_value} and~\ref{assum:three_dimensional} hold, then
the pmf at time $t+1$, $p_{\mathbf{Q}t+1}(\mathbf{q}_{t+1})$, can be computed from the pmf at time $t$ using the following
equation:
   \[ p_{\mathbf{Q}_{t+1}}(\mathbf{q}_{t+1}) = p_{\mathbf{Q}_t}(S^{-1}(T^{-1}(\mathbf{q}_{t+1}))) \]
\end{theorem}

\subsection{Algorithm Structure}

The \incx algorithm is schematically presented in~\Cref{fig:increx}, and its pseudo-code
is shown in \Cref{alg:increx}. 
In \Cref{fig:increx}, we list the object detectors used for the experiments in the paper, however \incx is agnostic to the detector, hence any object detector can be plugged in.

The saliency map in the first frame is  
computed using an \xai tool (in our experiments we used \drise).
The \emph{Explanation} procedure is used to extract a sufficient explanation from the saliency
map (see \Cref{subseq:exp} below).
After this bootstrapping, \incx computes saliency maps and sufficient explanations for subsequent frames
by performing the transformation described in \Cref{thm:translation} on the output of the previous frames.

As shown in \Cref{alg:increx}, in subsequent frames \incx maps the current objects to the objects that are
tracked in the previous frame using SORT~\cite{Bewley2016SimpleTracking}. 
Once the mapping is computed, the algorithm computes the linear transformation of the bounding boxes of the mapped objects by calling the Transform function 
(Line~\ref{alg:incx:transform}). This transformation is applied to the saliency map of the previous frame, using 
\Cref{thm:translation}. Note that if the new saliency map is scaled up, the gaps in the new pmf are filled using interpolation. 


The algorithm can compute explanations only for objects detected by the object tracker. An object might
leave the frame or remain undetected for a period of time and reappear later.
For efficiency, we define a \emph{timeout} that allows us to stop attempting to explain an object, which is no longer detectable.

\begin{algorithm}[t]
\caption{Pseudo-code of \incx}
\label{alg:increx}
\textbf{Input}: Video $v$, object detector $\mathcal{N}$, explainer $h$, tracker $\text{SORT}$\\
\textbf{Output}: List of explanations $xp$ and estimated saliency maps $s$
\begin{algorithmic}[1]
\STATE initialize bounding boxes and saliency map
\STATE initialize empty saliency list $s$ and explanation list $xp$
\FOR{each frame $f_t$ at time $t$ in $v$}
    \STATE $(x_t, y_t, w_t, h_t) \leftarrow \mathcal{N}(f_t)$
    \IF{$t = 0$} 
        \STATE $s_t \leftarrow h(f_t)$ \quad \COMMENT{Initial saliency map}
    \ELSE
        \STATE $s_t \leftarrow \text{Transform}(s_{t-1})$\COMMENT{track object(s) from previous frame}\label{alg:incx:transform}
    \ENDIF
    \STATE $xp_t \leftarrow \text{Explanation}(s_t, \mathcal{N}, f_t, t)$
    \STATE $xp \leftarrow xy \cup xp_t$
    \STATE $s \leftarrow s \cup s_t$
\ENDFOR
\STATE \textbf{return} $xp, s$
\end{algorithmic}
\end{algorithm}


\subsection{Sufficient explanations}\label{subseq:exp}

Thus far, we have discussed saliency maps. It is also useful to provide 
\emph{sufficient} explanations. We use the following definition, which is a standard
definition of causal explanations for image classifiers~\cite{CH24}.

\begin{definition}[\cite{CH24}]\label{defn:simple-exp}
An explanation is a \emph{minimal} subset of pixels of a
given input image that is \emph{sufficient} for the model to correctly detect the object,
with all other pixels of the image set to a baseline value (occluded).
\end{definition}
\noindent
Computing minimal and sufficient explanations is intractable~\cite{CH24}, so we relax the minimality
requirement. A \emph{sufficient explanation}
is a subset of pixels of the input image that is sufficient for the AI model to reproduce 
the original classification of the entire image. 
This is a useful sanity check of the saliency maps: the ``hot spots'' on the saliency map should be 
sufficient to trigger the original object detection. 
Sufficient explanations are computed by the \emph{explanation} procedure for both the initial and the
subsequent frames, as shown in \Cref{fig:increx}.
We note that, while the saliency maps for subsequent frames are approximate, the extracted explanations are
still correct, that is, sufficient for the original classification. 
They can, however, be larger than those computed from baseline saliency maps.

A brute-force extraction of sufficient explanations from saliency maps is inefficient, as it requires checking all subsets of pixels of the image. 
Our \emph{Explanation} procedure, illustrated in \Cref{fig:explanation_2} (see also the pseudo-code in the supplementary material) uses a binary
search on the saliency map of the current frame, as described below.

For the initial frame, the saliency map is divided into $l$ equal horizontal sections, as illustrated in \Cref{fig:explanation_1} (lateral view). The goal is to determine the smallest region that is sufficient for the model to make the same prediction. 
Therefore, we progressively occlude the pixels under a given dotted line. 
We then query the model with the partially occluded input to check whether the object is still being detected.

A naive approach would require querying the model at every possible dotted line, which is computationally expensive. Instead, we use binary search to efficiently locate a threshold (depicted by the red dotted line) that provides a sufficient, albeit not necessarily minimal, explanation. 
This approach works because the property of being sufficient is monotonic. In other words, starting from the top of the saliency map, 
as we reveal more of the image, the explanation becomes sufficient at some point and stays sufficient for all subsequent values.

Explanation discovery for subsequent frames is depicted in \Cref{fig:explanation_2}. Assuming temporal consistency in the saliency map, it is reasonable to expect that the explanation threshold in subsequent frames will be close to that of the previous frames. Therefore, we start our binary search 
around the previous threshold, with the number of divisions adjusted to a parameter $l_n$. Smaller values of $l_n$ improve performance but may result in explanations that contain more redundancy.

\begin{figure}[t]
\centering
    \begin{subfigure}[h]{0.22\textwidth}
        \centering
        \includegraphics[scale=0.55]{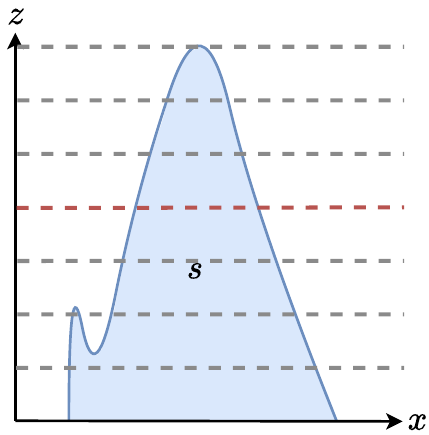}
        \caption{Explanation for the initial frame, with $l$ levels.}
        \label{fig:explanation_1}
    \end{subfigure}
    \hfill
    \begin{subfigure}[h]{0.22\textwidth}
        \centering
        \includegraphics[scale=0.6]{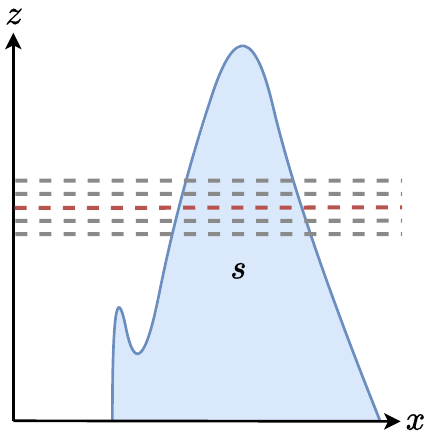}
        \caption{Explanation for subsequent frames, with $l_n$ levels.}
        \label{fig:explanation_2}
    \end{subfigure}
\caption{Visualization of the \emph{Explanation} procedure.}
\label{fig:explanation}
\end{figure}

\begin{table*}[t]
\centering
\renewcommand{\arraystretch}{1.2}
\setlength{\tabcolsep}{6pt}
\small
\begin{tabular}{l l l|c c c c c}
Dataset & Model & Explainer & Ins. ($\uparrow$) & Del. ($\downarrow$) & EPG ($\uparrow$) & EP ($\downarrow$) & Time(s) ($\downarrow$) \\
\midrule
\midrule
\multirow{6}{*}{BDD100K} 
    & \multirow{2}{*}{Faster R-CNN} 
        & D-RISE & $0.9459$ & $0.0233$ & $0.1974$ & $0.0125$ & $170.55$ \\ 
    &   & IncX & $0.9448$ & $0.0246$ & $0.2009$ & $0.0151$ & $1.06$ \\
    \cmidrule(lr){2-8}
    & \multirow{2}{*}{RT-DETR} 
        & D-RISE & $0.9321$ & $0.0510$ & $0.2542$ & $0.0387$ & $130.30$ \\
    &   & IncX & $0.9312$ & $0.0478$ & $0.2658$ & $0.0304$ & $2.54$ \\
    \cmidrule(lr){2-8}
    & \multirow{2}{*}{YOLO} 
        & D-RISE & $0.7879$ & $0.0326$ & $0.2771$ & $0.0816$ & $44.83$ \\
    &   & IncX & $0.8109$ & $0.0354$ & $0.3159$ & $0.0700$ & $1.22$ \\
\midrule
\multirow{6}{*}{KITTI} 
    & \multirow{2}{*}{Faster R-CNN} 
        & D-RISE & $0.9705$ & $0.0212$ & $0.0968$ & $0.0070$ & $129.06$ \\ 
    &   & IncX & $0.9646$ & $0.0214$ & $0.1384$ & $0.0074$ & $2.15$ \\
    \cmidrule(lr){2-8}
    & \multirow{2}{*}{RT-DETR} 
        & D-RISE & $0.9741$ & $0.0139$ & $0.0818$ & $0.0114$ & $114.09$ \\
    &   & IncX & $0.9760$ & $0.0157$ & $0.1071$ & $0.0131$ & $4.23$ \\
    \cmidrule(lr){2-8}
    & \multirow{2}{*}{YOLO} 
        & D-RISE & $0.9031$ & $0.0153$ & $0.1197$ & $0.0485$ & $40.32$ \\
    &   & IncX & $0.8970$ & $0.0158$ & $0.1330$ & $0.0521$ & $2.60$ \\
\midrule
\multirow{6}{*}{NuScenes} 
    & \multirow{2}{*}{Faster R-CNN} 
        & D-RISE & $0.9827$ & $0.0035$ & $0.0322$ & $0.0089$ & $203.56$ \\ 
    &   & IncX & $0.9771$ & $0.0034$ & $0.0366$ & $0.0118$ & $23.25$ \\
    \cmidrule(lr){2-8}
    & \multirow{2}{*}{RT-DETR} 
        & D-RISE & $0.9119$ & $0.0304$ & $0.1199$ & $0.0293$ & $225.65$ \\
    &   & IncX & $0.9011$ & $0.1182$ & $0.1201$ & $0.0272$ & $38.84$ \\
    \cmidrule(lr){2-8}
    & \multirow{2}{*}{YOLO} 
        & D-RISE & $0.7639$ & $0.0117$ & $0.1212$ & $0.0813$ & $95.45$ \\
    &   & IncX & $0.7777$ & $0.0132$ & $0.1202$ & $0.1178$ & $14.37$ \\
\midrule
\multirow{6}{*}{VIPER} 
    & \multirow{2}{*}{Faster R-CNN} 
        & D-RISE & $0.9417$ & $0.0452$ & $0.1705$ & $0.0177$ & $277.44$ \\ 
    &   & IncX & $0.9368$ & $0.0641$ & $0.1502$ & $0.0094$ & $22.91$ \\
    \cmidrule(lr){2-8}
    & \multirow{2}{*}{RT-DETR} 
        & D-RISE & $0.9465$ & $0.1420$ & $0.3373$ & $0.0617$ & $245.32$ \\
    &   & IncX & $0.9388$ & $0.1304$ & $0.4059$ & $0.0777$ & $27.91$ \\
    \cmidrule(lr){2-8}
    & \multirow{2}{*}{YOLO} 
        & D-RISE & $0.7128$ & $0.0555$ & $0.3284$ & $0.1280$ & $114.30$ \\
    &   & IncX & $0.7482$ & $0.0599$ & $0.4406$ & $0.1154$ & $9.80$ \\
\end{tabular}
\caption{Comparison of different models across multiple datasets using \drise and \incx as explainers. The arrows indicate whether higher ($\uparrow$) or lower ($\downarrow$) values of each metric are better.}
\label{tab:explainer}
\end{table*}
\section{Experimental Results}\label{sec:experimental_results}

\begin{table}[ht]
\centering
\small
\begin{tabular}{ l | l | c | c }
Dataset & Model & CC & SSIM \\
\midrule
\midrule
\multirow{3}{*}{BDD100K} & Faster R-CNN & $0.9395$ & $0.8599$ \\ 
& RT-DETR & $0.9648$ & $0.8383$ \\ 
& YOLO & $0.9307$ & $0.7447$ \\ 
\cmidrule{1-4}
\multirow{3}{*}{KITTI} & Faster R-CNN & $0.9071$ & $0.5832$ \\ 
& RT-DETR & $0.9118$ & $0.5625$ \\ 
& YOLO & $0.9308$ & $0.7471$ \\ 
\cmidrule{1-4}
\multirow{3}{*}{NUSCENES} & Faster R-CNN & $0.9433$ & $0.8284$ \\ 
& RT-DETR & $0.9028$ & $0.8023$ \\ 
& YOLO & $0.9221$ & $0.8494$ \\ 
\cmidrule{1-4}
\multirow{3}{*}{VIPER} & Faster R-CNN & $0.9147$ & $0.8308$ \\ 
& RT-DETR & $0.8661$ & $0.7124$ \\ 
& YOLO & $0.8203$ & $0.6464$ \\ 
\end{tabular}
\caption{Correlation Coefficient (CC) and Structural Similarity Index Measure (SSIM) scores used for comparing saliency maps across different models for each dataset.}
\label{tab:cc_ssim}
\end{table}

\begin{table}[ht]
\centering
\small
\begin{tabular}{ l | l | c | c }
Dataset & Model & JI & DC \\
\midrule
\midrule
\multirow{3}{*}{BDD100K} & Faster R-CNN & $0.3308$ & $0.4371$ \\ 
& RT-DETR & $0.5582$ & $0.6611$ \\ 
& YOLO & $0.6660$ & $0.7853$ \\ 
\cmidrule{1-4}
\multirow{3}{*}{KITTI} & Faster R-CNN & $0.4323$ & $0.5535$ \\ 
& RT-DETR & $0.5655$ & $0.6996$ \\ 
& YOLO & $0.6861$ & $0.7993$ \\ 
\cmidrule{1-4}
\multirow{3}{*}{NUSCENES} & Faster R-CNN & $0.5603$ & $0.6774$ \\ 
& RT-DETR & $0.5387$ & $0.6325$ \\ 
& YOLO & $0.7144$ & $0.7967$ \\ 
\cmidrule{1-4}
\multirow{3}{*}{VIPER} & Faster R-CNN & $0.4220$ & $0.5055$ \\ 
& RT-DETR & $0.4252$ & $0.5466$ \\ 
& YOLO & $0.6147$ & $0.7380$ \\ 
\end{tabular}
\caption{Jaccard Index (JI) and Dice Coefficient (DC) scores used for comparing explanations across different models for each dataset.}
\label{tab:ji_dc}
\end{table}

The COCO \cite{Lin2014MicrosoftContext} and VOC \cite{Everingham2010TheChallenge} datasets 
are typically used for evaluation of explainability in object detection~\cite{Kuroki2024FastDetection}. 
However, these datasets consist of independent images that lack temporal continuity, making them unsuitable for our focus on real-time video explanation for autonomous vehicles.
To address this limitation, we selected datasets specifically designed for autonomous driving, namely BDD100K~\cite{Yu_2020_CVPR}, KITTI~\cite{Geiger2013IJRR}, NuScenes~\cite{Caesar2020CVPR} and VIPER~\cite{Richter2017ICCV}, each of which contain hundreds of thousands of individual frames distributed across hundreds of videos. 
Given the computational expense of explanation methods such as \drise, processing these datasets in full would be impractical. 
To ensure feasible benchmarking, we randomly selected a representative subset from each dataset.  From KITTI, BDD100K, and VIPER, we chose five videos each and extracted 100 frames per video. For KITTI, this corresponds to 10 seconds per video at 10 FPS. For NuScenes, we used the entire mini dataset, consisting of 10 videos of 404 frames.


We evaluated \incx on the following object detectors: {\sc yolo}v10-n, 
{\sc rt-detr}-l, and the PyTorch implementation of \fastrcnn. 
\drise was used to compute the initial saliency maps. 
\incx makes no assumptions about how the initial saliency map is produced: \drise is used as it works naturally with 
the different object detectors that we examine and is considered the state-of-the-art explainability tool for object detectors. 
For comparison we also computed the saliency maps independently using \drise for all frames. 
We executed \drise with $1000$ mutants and a mask resolution of $(4,4)$. 
All experiments were conducted using the NVIDIA A100 Tensor Core GPU.


For a one-to-one comparison between each saliency map generated by \drise and \incx, we computed a set of metrics commonly used in the field of Explainable AI. We describe these metrics below.

The \textbf{Normalized Insertion/Deletion curves} metric starts with a blank image, 
adding pixels based on the saliency map, and computes the area under the curve (AUC) 
after measuring detector confidence. Conversely, the deletion process starts with the full image and 
gradually removes pixels according to the saliency map. For object detectors, 
we adjust this metric by multiplying the score by the Intersection over Union (IoU) of the original bounding box. This adjustment accounts for the fact that traditional insertion and deletion curves assess whether an object is detected, but do not consider its precise location, an essential factor for object detection tasks.
To account for low confidence producing low insertion and deletion values, 
we normalize the curve by dividing all values by the original image's softmax classification. 

The \textbf{Energy-based Pointing Game (EPG)} metric quantifies the proportion of the 
saliency map's total energy that is contained within the bounding box \cite{Wang2019Score-CAM:Networks}. 
This metric is essential because an accurate saliency map should predominantly cover the bounding box to 
correctly represent the object of interest. If the saliency map is focused outside of the bounding box, it may indicate the detection of an object that does not correspond to the one within the current bounding box. 

The \textbf{Explanation Proportion (EP)} metric measures explanation quality by 
computing the ratio of the explanation to the whole image, hence favoring compact explanations that minimize unnecessary information. We define EP as follows: for a given boolean-valued mask $m$ of an image, 
where $\top$ denotes the pixels included in the explanation, the EP is computed as
\begin{equation}
\text{EP} = \frac{\sum_{(i,j) \in \top} m_{(i,j)}}{\sum_{(i,j) \in \top} m_{(i,j)} + \sum_{(i,j) \notin \top} m_{(i,j)}}
\end{equation}

As summarized in \Cref{tab:explainer}, the insertion values for \incx are consistently within 5\% of those observed for \drise, with all other metrics displaying similar trends. Moreover, the running time of \incx 
is two orders of magnitude less than that of \drise in most cases, 
showing that \incx is suitable for real-time scenarios.

The quality of an approximate saliency map computed by \incx is evaluated by measuring its deviation it from the
baseline saliency map, computed by \drise directly from the image. 
We compute this deviation by using standard metrics: Pearson Correlation Coefficient (CC), Structural Similarity Index (SSIM)~\cite{Wang2004ImageSimilarity} (\Cref{tab:cc_ssim}), Dice Coefficient (DC)~\cite{Dic45}, and Jaccard Index (IoU) (\Cref{tab:ji_dc}). 
The results in both tables indicate a strong similarity between the saliency maps generated by \incx and the baseline maps produced by \drise.

\subparagraph*{Limitations}
\incx assumes that the object it is applied to is rigid and does not rotate.
Our experimental results show that this assumption is reasonable in practice for autonomous driving.
Another implicit assumption is that consecutive frames produce similar saliency maps, which is
supported by the laws of physics: objects tend not to disappear or (re-)materialize.
\section{Conclusions}
In this paper we have presented the algorithm \incx for real-time explanations of object detectors. The algorithm
incurs a negligible overhead on the object detector and is two orders of magnitude faster than the state-of-the-art. Our method does not lead to inferior quality saliency
maps, despite its assumptions. The standard quality metrics for assessing \xai algorithms show that
\incx is comparable to \drise, the state-of-the-art in black-box \xai for object detectors.
Evaluation across multiple standard datasets supports our claim that \incx is suitable for
autonomous driving applications.

\newpage
\bibliographystyle{ieeenat_fullname}
\bibliography{references,all}

\end{document}


\title{Supplementary Material for the paper\\
``Real-Time Incremental Explanations for Object Detectors\\ in Autonomous Driving''}
\author{
    Santiago Calder\'{o}n-Pe\~{n}a,
    Hana Chockler,
    David A. Kelly \\
    King's College London \\
    London, United Kingdom \\
    {\tt\small \{santiago.calderon, hana.chockler, david.a.kelly\}@kcl.ac.uk}
}
\author{}

\maketitle

\section{Proofs}

For the ease of reading, we repeat the statement of the main theorem and the relevant definitions here.


Transformations of probability mass functions (pmfs) for random variables can also be extended to bivariate random variables, as shown in~\cref{thm:prob_conv}.~\cite{Hogg2019IntroductionStatistics}.

\begin{theorem}
\label{thm:prob_conv}
Let $p_{X_1,X_2}(x_1, x_2)$ be the joint pmf of two discrete random variables $X_1$ and $X_2$. Let $x_1=w_1(y_1, y_2)$ and $x_2=w_2(y_1, y_2)$ define a one-to-one transformation. The joint pmf of the two new random variables $Y_1=u_1(X_1, x_2)$ and $Y_2=u_2(X_1, X_2)$ with support $\mathcal{T}$ is given by

 \begin{equation}
p_{Y_,Y_2}(y_{1},y_{2}) = 
     \begin{cases} 
     \!\begin{aligned}
     p_{X_{1},X_2}[w_{1}(y_{1}, y_{2}),\\ w_2(y_1, y_2)]
     \end{aligned} & (y_1, y_2) \in \mathcal{T} \\
     0              & \text{elsewhere}
\end{cases}
\end{equation}

\end{theorem}

The projection from 3D world coordinates to camera image coordinates can be effectively modeled using the pinhole camera approximation~\cite[p. 129]{Forsyth2002ComputerVision}.

\begin{lemma}[3D world coordinates to 2D screen (camera image) coordinates]
\label{lem:projection} 
Let \(\alpha, \beta \in \mathbb{R}\) denote the intrinsic magnification factors of the camera. The transformation from 3D world coordinates \((x, y, z)\) to 2D screen coordinates \((u, v)\) is given by:

\[ \begin{bmatrix}
u \\ v
\end{bmatrix} = \frac{1}{z} \begin{bmatrix}
\alpha & 0 & O_x \\
0 & \beta & O_y
\end{bmatrix} \begin{bmatrix}
x \\
y \\
z
\end{bmatrix}= \frac{1}{z} \mathcal{K} \begin{bmatrix}
x \\
y \\
z
\end{bmatrix} \]

where \(\mathcal{K}\) represents the intrinsic parameters of the camera, and \((O_x, O_y)\) represents the coordinates at the upper-left corner of the image.

\end{lemma}

\begin{definition}[Center Function]
\label{def:center}
The center function, denoted as $\mathbf{C}: 2^{\mathbb{R}^n} \to \mathbb{R}^n$, maps a set of points to the center of the bounding box that encloses them. For a set $S$ consisting of points $\mathbf{p} = (p_{1}, p_{2}, \ldots, p_{n}) \in \mathbb{R}^n$, the function is defined as follows:

\[
\mathbf{C}(S) = \begin{bmatrix}
\frac{\max(p_{1}) + \min(p_{1})}{2} \\
\frac{\max(p_{2}) + \min(p_{2})}{2} \\
\vdots \\
\frac{\max(p_{n}) + \min(p_{n})}{2}
\end{bmatrix}
\]

where $max(p_j)$ and $min(p_j)$ denote the maximum and minimum values, respectively, of the $j$-th coordinate among all points in $S$.
\end{definition}

\begin{definition}[Scaling and Translation]
\label{def:sca_trans}
Let \( A = \{(u, v) \in \mathbb{R}^2 \} \) represent a set of points defining any area in the image space.

\noindent
\textbf{Scaling} is a linear transformation $S: A \to \mathbb{R}^2$, defined as \( S(\mathbf{p}) = \gamma(\mathbf{p} - \mathbf{C}(A)) \), where \( \mathbf{p} \in A \) and \( \gamma \in \mathbb{R}^{2 \times 2} \) is a diagonal matrix. 

\noindent
\textbf{Translation} is a linear transformation $T: A \to \mathbb{R}^2$, defined as $T(\mathbf{p}) = \mathbf{p} + \mu$, where $\mathbf{p}\in A$. 
\end{definition}

Let \( \mathbf{Q}_t = (Q_{1,t}, Q_{2,t}) \) be a vector in the image space at time $t$, where $Q_{1,t}$ and $Q_{2,t}$ represent random variables along the $u$-axis and the $v$-axis, respectively. The pmf of $\mathbf{Q}_t$ is \( p_{\mathbf{Q}_t}(\mathbf{q}_{t}) \), which is derived from the normalized saliency map.

\begin{assumption}[Expected Value and Bounding Box Center Relation] \label{assum:expected_value}

Let \( V_t \) denote the set of points of the three-dimensional object, and let \( \delta_t \) represent the difference between the expected value of the probability mass function (pmf) and the center of the bounding box. The expected value of \( \mathbf{Q}_t \) is related to the center of the connected object as follows:

\[
\mathbb{E}[\mathbf{Q}_t] = \mathbf{C}(\mathcal{K}V_t) + \delta_t
\]

\end{assumption}

\begin{figure*}[t]
\begin{equation}\label{equ:relation}
\begin{bmatrix}
u_{t+1} \\
v_{t+1}
\end{bmatrix} =
\frac{1}{m_z(z_t + \Delta z)}
\begin{bmatrix}
\alpha & 0 & O_x & 0 \\
0 & \beta & O_y & 0
\end{bmatrix}
\begin{bmatrix}
m_x & 0 & 0 & 0 \\
0 & m_y & 0 & 0 \\
0 & 0 & m_z & 0 \\
0 & 0 & 0 & 1
\end{bmatrix}
\begin{bmatrix}
    \mathbf{I} & \mathbf{t} \\
    \mathbf{0}^T & 1
\end{bmatrix}
\begin{bmatrix}
x_t \\
y_t \\
z_t \\
1
\end{bmatrix}
\end{equation}
\end{figure*}

\begin{assumption}[Constrained Affine Transformation] \label{assum:three_dimensional}

Any point of an object at time \( t+1 \) is related to the same object at time \( t \) by an affine transformation given by:

\[
\begin{bmatrix}
    x_{t+1} \\
    y_{t+1} \\
    z_{t+1} \\
    1
\end{bmatrix}
=
\begin{bmatrix}
    m_x & 0 & 0 & 0 \\
    0 & m_y & 0 & 0 \\
    0 & 0 & m_z & 0 \\
    0 & 0 & 0 & 1
\end{bmatrix}
\begin{bmatrix}
    \mathbf{I} & \mathbf{t} \\
    \mathbf{0}^T & 1
\end{bmatrix}
\begin{bmatrix}
    x_t \\
    y_t \\
    z_t \\
    1
\end{bmatrix}
\]

where \( \mathbf{t} \in \mathbb{R}^3 \) is an arbitrary translation vector, \( \mathbf{I} \in \mathbb{R}^{3 \times 3} \) is the identity matrix, and \( m_x \), \( m_y \), and \( m_z \in \mathbb{R} \) are scaling factors.

\end{assumption}

Let \( S \) be the scaling function, \( T \)  the translation function, \( \mathbf{q}_{t+1} \) a coordinate vector in the image space at time \( t+1 \), and \( \mathbf{q}_{t} \) a coordinate vector at time \( t \). 
The following theorem describes the relation between the pmf at time $t$ and at time $t+1$ and is instrumental to the \incx algorithm.
\begin{theorem}\label{thm:translation}
If Assumptions~\ref{assum:expected_value} and~\ref{assum:three_dimensional} hold, then
the pmf at time $t+1$, $p_{\mathbf{Q}t+1}(\mathbf{q}_{t+1})$, can be computed from the pmf at time $t$ using the following
equation:
   \[ p_{\mathbf{Q}_{t+1}}(\mathbf{q}_{t+1}) = p_{\mathbf{Q}_t}(S^{-1}(T^{-1}(\mathbf{q}_{t+1}))) \]
\end{theorem}

\begin{proof}
From Assumption \ref{assum:three_dimensional} and \cref{lem:projection}, we start by analyzing how the projection of a point \((x_t, y_t, z_t)\) at time \(t\) transforms into \((u_{t+1}, v_{t+1})\) at time \(t+1\). Using the projection model, the relationship is given by~\cref{equ:relation}.

Let $\mathbf{t} = \begin{bmatrix} \Delta x & \Delta y & \Delta z \end{bmatrix}^{T}$ be the translation vector from an object at time $t$ to the same object at time $t+1$. Simplifying this matrix multiplication and rearranging yields \cref{eq:u_tp1_1}.


\begin{equation}\label{eq:u_tp1_1}
\small
\begin{aligned}
\begin{bmatrix} 
u_{t+1} \\ 
v_{t+1} 
\end{bmatrix} &= \frac{1}{m_z \cdot (z_t + \Delta z)} 
\begin{bmatrix} 
m_x & 0 \\ 
0 & m_y 
\end{bmatrix} 
\begin{bmatrix} 
\alpha \cdot x_t \\ 
\beta \cdot y_t 
\end{bmatrix} \\
&\quad + \frac{1}{m_z \cdot (z_t + \Delta z)} 
\begin{bmatrix} 
\alpha \cdot m_x \cdot \Delta x + m_z \cdot O_x \cdot (z_t + \Delta z) \\ 
\beta \cdot m_y \cdot \Delta y + m_z \cdot O_y \cdot (z_t + \Delta z) 
\end{bmatrix}
\end{aligned}
\end{equation}

From \cref{lem:projection} we can also obtain the following expression:

\begin{equation}
\label{eq:u_tp_1}
\begin{bmatrix}
\alpha \cdot x_t \\
\beta \cdot y_t
\end{bmatrix} = z_t
\begin{bmatrix}
u_t \\
v_t
\end{bmatrix}
- 
\begin{bmatrix}
O_x \cdot z_t \\
O_y \cdot z_t
\end{bmatrix}
\end{equation}

\begin{figure*}
\begin{equation}\label{eq:u_tp1_2}
\begin{bmatrix}
u_{t+1} \\
v_{t+1}
\end{bmatrix} = \frac{z_t}{m_z \cdot (z_t + \Delta z)}
\begin{bmatrix}
m_x & 0 \\
0 & m_y
\end{bmatrix}
\begin{bmatrix}
u_t \\
v_t
\end{bmatrix} + \frac{1}{m_z \cdot (z_t + \Delta z)}
\begin{bmatrix}
\alpha \cdot m_x \cdot \Delta x + m_z \cdot O_x \cdot (z_t + \Delta z) - O_x \cdot z_t \\
\beta \cdot m_y \cdot \Delta y + m_z \cdot O_y \cdot (z_t + \Delta z) - O_y \cdot z_t
\end{bmatrix}
\end{equation}
\end{figure*}

Substituting \cref{eq:u_tp_1} into \cref{eq:u_tp1_1} yields \cref{eq:u_tp1_2}. Note that all values are constant except for $\begin{bmatrix} u_{t+1} & v_{t+1} \end{bmatrix}^{T}$ and $\begin{bmatrix} u_{t} & v_{t} \end{bmatrix}^{T}$. Consequently, for simplicity, we denote this expression as follows:

\begin{equation}\label{eq:u_tpm1}
\begin{bmatrix}
u_{t+1} \\
v_{t+1}
\end{bmatrix} = K_1
\begin{bmatrix}
u_t \\
v_t
\end{bmatrix} + K_2
\end{equation}

This relationship holds for any point in the image. Therefore, it extends to the maximum and minimum values of the coordinates, as stated in ~\cref{equ:min}
and~\cref{equ:max}.

\begin{equation}\label{equ:min}
\begin{bmatrix}
\max(u_{t+1}) \\
\max(v_{t+1})
\end{bmatrix} = K_1
\begin{bmatrix}
\max(u_t) \\
\max(v_t)
\end{bmatrix} + K_2
\end{equation}

\begin{equation}\label{equ:max}
\begin{bmatrix}
\min(u_{t+1}) \\
\min(v_{t+1})
\end{bmatrix} = K_1
\begin{bmatrix}
\min(u_t) \\
\min(v_t)
\end{bmatrix} + K_2
\end{equation}

As a result, $K_1$ can be derived as:

\begin{equation}\label{eq:k_1}
K_1 = \begin{bmatrix}
\frac{\max(u_{t+1}) - \min(u_{t+1})}{\max(u_t) - \min(u_t)} & 0 \\
0 & \frac{\max(v_{t+1}) - \min(v_{t+1})}{\max(v_t) - \min(v_t)}
\end{bmatrix}
\end{equation}

Consider the projected object, denoted by the set of points \(A_t\), where:

\[
A_t = \{ (u,v) \in \mathbb{R}^2 \}
\]

We can apply a similar procedure to the center of the bounding box at times \(t\) and \(t+1\), resulting in:

\[
\begin{bmatrix}
\mathbf{C}(A_{t+1}) \\
\mathbf{C}(A_{t+1})
\end{bmatrix} = K_1
\begin{bmatrix}
\mathbf{C}(A_t) \\
\mathbf{C}(A_t)
\end{bmatrix} + K_2
\]

The previous equation allows us to find $K_2$ in terms of the center of the bounding box, as follows:

\begin{equation}\label{eq:k_2}
K_2 =
\begin{bmatrix}
\mathbf{C}(A_{t+1}) \\
\mathbf{C}(A_{t+1})
\end{bmatrix} - K_1
\begin{bmatrix}
\mathbf{C}(A_t) \\
\mathbf{C}(A_t)
\end{bmatrix}
\end{equation}

Substituting the expressions from \cref{eq:k_1} and \cref{eq:k_2} into \cref{eq:u_tpm1}, we obtain \cref{eq:u_tp_final}. This demonstrates how the current bounding box is transformed into the new bounding box.

\begin{figure*}
\begin{equation}
\label{eq:u_tp_final}
\begin{bmatrix}
u_{t+1} \\
v_{t+1}
\end{bmatrix} = \begin{bmatrix}
\frac{\max(u_{t+1}) - \min(u_{t+1})}{\max(u_t) - \min(u_t)} & 0 \\
0 & \frac{\max(v_{t+1}) - \min(v_{t+1})}{\max(v_t) - \min(v_t)}
\end{bmatrix}
\left(
\begin{bmatrix}
u_t \\
v_t
\end{bmatrix} - \begin{bmatrix}
c(A_t) \\
c(A_t)
\end{bmatrix}
\right) +
\begin{bmatrix}
c(A_{t+1}) \\
c(A_{t+1})
\end{bmatrix}
\end{equation}
\end{figure*}

Equation~\eqref{eq:u_tp_final} is crucial for our algorithm as it allows us to express any change in the three-dimensional object, 
only as a change in elements that can be known via an object detector, i.e., the bounding box corners and its center.

Following Definition \ref{def:sca_trans}, ~\cref{eq:u_tp_final} can be expressed in terms of the Scaling and Translation transformation as follows:

\begin{equation}
\label{eq:u_tp1}
\mathbf{U}_{t+1} = T(S(\mathbf{U}_t)),
\end{equation}

where 
\begin{equation}
\mathbf{U}_{t} = \begin{bmatrix}
u_t \\
v_t
\end{bmatrix}
\end{equation}

This first part of the proof essentially shows that any projected point of a three-dimensional object that has only undergone a translation and scaling from the previous frame can be represented in the current frame just as a Scaling and Translation, which are common transformations in the computer vision domain.

In the next part of the proof, we demonstrate that this transformation further expands to a saliency map that is connected to this object.

From the first assumption of the theorem, we have:

\[
\mathbb{E}[\mathbf{Q}_t] = \mathbf{C}(\mathcal{K}V_t) + \delta_t
\]

This indicates that the expected value of the random vector \(\mathbf{Q}_t\) is a projection of a point in the 3D space onto the camera plane, with an additional offset \(\delta_t\). Consequently, this point belongs to the projected object and we can use~\cref{eq:u_tp1}, obtaining

\[
\mathbb{E}[\mathbf{Q}_{t+1}] = T(S(\mathbb{E}[\mathbf{Q}_{t}]))
\]

Since \( T \) and \( S \) are linear functions, and noting that the entire pmf is scaled and translated in the same manner as the expected value, we have:

\[
\mathbf{Q}_{t+1} = T(S(\mathbf{Q}_t))
\]

This implies that:

\[
\mathbf{Q}_t = S^{-1}(T^{-1}(\mathbf{Q}_{t+1}))
\]

To obtain the joint distribution of \(\mathbf{Q}_{t+1}\), we can leverage Theorem \ref{thm:prob_conv}, obtaining the following expression

\[ p_{\mathbf{Q}_{t+1}}(\mathbf{q}_{t+1}) = p_{\mathbf{Q}_t}(S^{-1}(T^{-1}(\mathbf{q}_{t+1}))) \]

This completes the proof.

\end{proof}

\section{Explanation Algorithm}

The \textbf{Explanation algorithm} is guaranteed to find a sufficient explanation for the first frame. However, if no explanation is found within the defined margin for the subsequent frames, the algorithm defaults to using the explanation area from the previous frame.

\Cref{alg:explanation} provides a pseudo-code of this algorithm. For the first frame, the search space \(ss\) spans from the minimum to maximum values of the saliency map \(s\), divided into $l$ equally spaced levels. For subsequent frames, an additional parameter \(\delta\) defines the margin around the previous \(\text{threshold}_{t-1}\), adjusting the search space boundaries accordingly and setting the number of divisions to $l_n$. A binary search is performed within the defined search space for each case to determine the explanation.

\begin{algorithm}[ht]
\caption{Explanation Algorithm}
\label{alg:explanation}
\textbf{Input}: Saliency map \(s\), object detector \(g\), image \(I\), time step \(t\), margin \(\delta\), number of divisions $l$ (for $t=0$) and $l_n$ (for \(t > 0\))\\
\textbf{Output}: Explanation \(xp\)
\begin{algorithmic}[1]
\IF {$t == 0$}
    \STATE \(\text{min} \leftarrow \min(s)\)
    \STATE \(\text{max} \leftarrow \max(s)\)
    \STATE \( ss \leftarrow \text{Linspace}(\text{min}, \text{max}, l)\)
\ELSE
    \STATE \(\text{max} \leftarrow \text{threshold}_{t-1} \cdot (1+\delta)\)
    \STATE \(\text{min} \leftarrow \text{threshold}_{t-1} \cdot (1-\delta)\)
    \STATE \( ss \leftarrow \text{Linspace}(\text{min}, \text{max}, l_n)\)
\ENDIF
\STATE \(xp, \text{threshold}_{t-1} \leftarrow \text{BinarySearch}(s, g, I, ss)\)
\STATE \textbf{return} \(xp\)
\end{algorithmic}
\end{algorithm}

\section{Metrics}


In this section we provide the details of the various metrics used in the paper.

\paragraph{Normalized Insertion:} The non-normalized version of this metric is extensively used in explainable AI and has been analyzed across various explainability tools \cite{Petsiuk2020Black-boxMaps,Kuroki2024FastDetection,petsiuk2018rise}. In image classification tasks, the procedure starts with a black image, progressively adding pixels from the saliency map. The metric involves computing the Area Under the Curve (AUC) of the plot generated by this process.

The principle underlying this approach is that incremental addition of features from the saliency map allows for assessing their contribution to model performance. Higher insertion values indicate a substantial positive impact on the model's prediction accuracy when features are reintroduced. Elevated insertion values reflect that the feature significantly enhances the model's performance, highlighting its critical role and relevance. This improves model interpretability by identifying influential features and increases stakeholder trust in the model’s predictions.

For object detection, it is essential to incorporate the bounding box, as multiple instances of the same class may be present. Consequently, the metric is adjusted by multiplying the confidence score by the Intersection over Union (IoU) of the original bounding box. This adjustment ensures that the metric accurately reflects the confidence for each detected object, with a separate saliency map generated for each object.

While this metric is useful, it has major limitations. Specifically, if the initial confidence in the detected object is low, the insertion score tends to be low, making the score highly dependent on the specific data and model. To address this issue, we normalize the insertion value by dividing the entire graph by the confidence level of the detected object when presented with the complete image. This curve starts at zero, as it begins with a black image, and the model's confidence increases as pixels are progressively added. Additionally, unlike the typical insertion curve, this normalized curve may sometimes exceed a value of 1, indicating scenarios where the model exhibits greater confidence in its output with a subset of pixels than with the entire set.


\paragraph{Normalized Deletion:} Deletion metrics are typically computed alongside insertion metrics and can be viewed as complementary to them. The process begins with the full image, and pixels are sequentially removed according to the saliency map. As with insertion, the Area Under the Curve (AUC) is calculated. Ideally, the deletion score should be lower because removing pixels from crucial regions (as indicated by high values in the saliency map) is expected to significantly decrease the model's confidence in its classification.

To ensure the metric remains relevant to the object of interest, we account for the bounding box by scaling the results based on the Intersection over Union (IoU) of the original bounding box. Furthermore, we normalize the result as we did with the normalized insertion metric. This curve typically starts with high confidence and decreases as pixels are removed.


\paragraph{Energy-based Pointing Game (EPG):} The Energy-based Pointing Game (EPG) metric evaluates the alignment of the saliency map with the target bounding box \cite{Wang2019Score-CAM:Networks}. It quantifies the proportion of the saliency map's total energy that is contained within the bounding box. This is crucial because the saliency map should primarily cover the bounding box to accurately reflect the object of interest. For example, a saliency map might produce a high classification score for the desired class, but if it is entirely outside the bounding box, this suggests that the saliency map is not accurately targeting the intended object and may be focusing on a different object.

The EPG metric is defined for a saliency map $h$ as:

$$
EPG = \frac{\sum_{(i,j) \in \text{BBox}} h_{(i,j)}}{\sum_{(i,j) \in \text{BBox}} h_{(i,j)} + \sum_{(i,j) \notin \text{BBox}} h_{(i,j)}}
$$

\paragraph{Explanation Proportion (EP):} In addition to evaluating the saliency maps, it is crucial to assess the effectiveness of the explanations themselves. Previous research indicates that an ideal explanation should be compact \cite{Chockler2023MultipleClassifiers}. A compact explanation implies that the saliency map exhibits a sharp drop in relevance outside the selected pixels, suggesting that these pixels alone are sufficient for classification.

\section{Code and Datasets}

The code for \incx and datasets can be found at the following link:

\url{https://figshare.com/s/adb210fa4bfc67f142fa}

\noindent Instructions for running IncX and replicating the results from all the experiments detailed in the paper are provided in the \texttt{README.md} file. The specific subsets of the datasets used in the experiments are available in the \texttt{dataset} subfolder.

\section{Docker Image}

To view the functionality of IncX on your local machine, follow these steps:

\begin{enumerate}
    \item Pull the Docker image using the following command:
\begin{verbatim}
docker pull incx/incx-web-application
\end{verbatim}
    \item Run a container from the pulled image.
    \item Open your web browser and navigate to\\ \url{http://127.0.0.1:8000/} to access the application.
\end{enumerate}

\bibliographystyle{plainnat}
\bibliography{references,all}